# Structured Knowledge Representation for Image Retrieval

**Eugenio Di Sciascio**  DISCIASCIO@POLIBA.IT
**Francesco M. Donini**  DONINI@POLIBA.IT
**Marina Mongiello**  MONGIELLO@POLIBA.IT
*Dipartimento di Elettrotecnica ed Elettronica, Politecnico di Bari*
*Via Re David, 200 – 70125 BARI Italy*

## Abstract

We propose a structured approach to the problem of retrieval of images by content and present a description logic that has been devised for the semantic indexing and retrieval of images containing complex objects.

As other approaches do, we start from low-level features extracted with image analysis to detect and characterize regions in an image. However, in contrast with feature-based approaches, we provide a syntax to describe segmented regions as basic objects and complex objects as compositions of basic ones. Then we introduce a companion extensional semantics for defining reasoning services, such as retrieval, classification, and subsumption. These services can be used for both exact and approximate matching, using similarity measures.

Using our logical approach as a formal specification, we implemented a complete client-server image retrieval system, which allows a user to pose both queries by sketch and queries by example. A set of experiments has been carried out on a testbed of images to assess the retrieval capabilities of the system in comparison with expert users ranking. Results are presented adopting a well-established measure of quality borrowed from textual information retrieval.

## 1. Introduction

Image retrieval is the problem of selecting, from a repository of images, those images fulfilling to the maximum extent some criterion specified by an end user. In this paper, we concentrate on *content-based* image retrieval, in which criteria express properties of the appearance of the image itself, *i.e.*, on its pictorial characteristics.

Most of the research in this field has till now concentrated in devising suitable techniques for extracting relevant cues with the aid of image analysis algorithms. Current systems result effective when the specified properties are so-called *low-level* characteristics, such as color distribution, or texture. For example, systems such as IBM's QBIC[1] can easily retrieve, among others, stamps containing the picture of a brown horse in a green field, when asked to retrieve images of stamps with brown central area over a greenish background.

Nevertheless, present systems fail at treating correctly *high-level* characteristics of an image — such as, "retrieve stamps with a galloping horse". First of all, most systems cannot even allow the user to specify such queries, because they lack a language for expressing *high-level* features. Usually, this is overcome with the help of examples: "retrieve images similar to *this* one". However, examples are quite ambiguous to interpret: which are the features in the example that should appear in retrieved images? This ambiguity produces a lot of "false positives", as any one can experience.

---

1. See *e.g.*, `http://wwwqbic.almaden.ibm.com/cgi-bin/stamps-demo`

Even if relevant features are pointed out in the example, the system cannot tell whether what is pointed out is the color distribution, or its interpretation — after all, a galloping brown horse produces a color distribution which is more similar to a running brown fox than to a galloping *white* horse. In this aspect, image retrieval faces the same problems of *object recognition*, which is a central problem in robotics and artificial vision. The only effective solution overcoming this problem is to associate to a query some significant keywords, which should match keywords attached in some way to images in the repository. Here ambiguities in image understanding are just transferred to text understanding, as now a brown portrait of Crazy Horse — the famous Indian chief — could be considered relevant.

Resorting to human experts to specify the expected output of a retrieval algorithm can, in our opinion, only worsen these ambiguities, since it makes the correctness of an approach to depend from a subjective perception of what an image retrieval system should do. What is needed is a formal, high-level specification of the image retrieval task. This need motivates the research we report here.

### 1.1 Contributions of the Paper

We approach the problem of image retrieval from a knowledge representation perspective, and in particular, we refer to a framework already successfully applied by Woods and Schmolze (1992) to conceptual modeling and semantic data models in databases (Calvanese, Lenzerini, & Nardi, 1998). We consider image retrieval as a knowledge representation problem, in which we can distinguish the following aspects:

**Interface:** the user is given a simple visual language to specify (by sketch or by example) a geometric composition of basic shapes, which we call description. The composite shape description intuitively stands for a set of images (all containing the given shapes in their relative positions); it can be used either as a query, or as an index for a relevant class of images, to be given some meaningful name.

**Syntax and semantics:** the system has an internal syntax to represent the user's queries and descriptions, and the syntax is given an extensional semantics in terms of sets of retrievable images. In contrast with existing image retrieval systems, our semantics is compositional, in the sense that adding details to the sketch may only restrict the set of retrievable images. Syntax and semantics constitute a Semantic Data Model, in which the relative position, orientation and size of each shape component are given an explicit notation through a geometric transformation. The extensional semantics allows us to define a hierarchy of composite shape descriptions, based on set containment between interpretations of descriptions. Coherently, the recognition of a shape description in an image is defined as an interpretation satisfying the description.

**Algorithms and complexity:** based on the semantics, we prove that subsumption between descriptions can be carried out in terms of recognition. Then we devise exact and approximate algorithms for composite shapes recognition in an image, which are correct with respect to the semantics. Ideally, if the computational complexity of the problem of retrieval was known, the algorithms should also be optimal with reference to the computational complexity of the problems. Presently, we solved the problem for exact retrieval, and propose an algorithm for approximate retrieval which, although probably non-optimal, is correct.

**Experiments:** while the study of the complexity of the problem is ongoing, the syntax, semantics, and sub-optimal algorithms obtained so far are already sufficient to provide the formal specification of a prototype system for the experimental verification of our approach. The prototype has been used to carry out a set of experiments on a test database of images, which allowed us to verify the effectiveness of the proposed approach in comparison with expert users ranking.

We believe that a knowledge representation approach brings several benefits to research in image retrieval. First of all, it separates the problem of finding an intuitive semantics for query languages in image retrieval from the problem of implementing a correct algorithm for a given semantics. Secondly, once the problem of image retrieval is semantically formalized, results and techniques from Computational Geometry can be exploited in assessing the computational complexity of the formalized retrieval problem, and in devising efficient algorithms, mostly for the *approximate* image retrieval problem. This is very much in the same spirit as finite model theory has been used in the study of complexity of query answering for relational databases (Chandra & Harel, 1980). Third, our language borrows from object modeling in Computer Graphics the hierarchical organization of classes of images (Foley, van Dam, Feiner, & Hughes, 1996). This, in addition to an interpretation of composite shapes which one can immediately visualize, opens our logical approach to retrieval of images of 3D-objects constructed in a geometric language (Paquet & Rioux, 1998), which is still to be explored. Fourth, our logical formalization, although simple, allows for extensions which are natural in logic, such as disjunction (OR) of components. Although alternative components of a complex shape are difficult to be shown in a sketch, they could be used to specify moving (*i.e.*, non-rigid) parts of a composite shape. This exemplifies how our logical approach can shed light to extensions of our syntax suitable for, *e.g.*, video sequence retrieval.

### 1.2 Outline of the Paper

The rest of the paper is organized as follows. In the next section, we review related work on image retrieval. In Section 3 we describe our formal language, first its syntax, then its semantics, and we start proving some basic properties. In the following section, we analyze the reasoning problems and the semantic relations among them, and we devise algorithms that can solve them. Then in Section 5 we illustrate the architecture of our system and propose some examples pointing out distinguishing aspects of our approach. In Section 6 we present a set of experiments to assess retrieval capabilities of the system. Last section draws the conclusions and proposes directions for future work.

## 2. Related Work

Content-Based Image Retrieval (CBIR) has recently become a widely investigated research area. Several systems and approaches have been proposed; here we briefly report on some significant examples and categorize them in three main research directions.

### 2.1 Feature-based Approaches

Largest part of research on CBIR has focused on low-level features such as color, texture, shape, which can be extracted using image processing algorithms and used to characterize an image in some feature space for subsequent indexing and similarity retrieval. In this way the problem of retrieving images with homogeneous content is substituted with the problem of retrieving images visually close to a target one (Hirata & Kato, 1992; Niblak et al., 1993; Picard & Kabir, 1993; Jacobs, Finkelstein, & Salesin, 1995; Flickner et al., 1995; Bach, Fuller, Gupta, Hampapur, Horowitz, Humphrey, Jain, & Shu, 1996; Celentano & Di Sciascio, 1998; Cox, Miller, Minka, & Papathomas, 2000; Gevers & Smeulders, 2000).

Among the various projects, particularly interesting is the QBIC system (Niblak et al., 1993; Flickner et al., 1995), often cited as the ancestor of all other CBIR systems, which allows queries to be performed on shape, texture, color, by example and by sketch using as target media both images and shots within videos. The system is currently embedded as a tool in a commercial product, ULTIMEDIA MANAGER. Later versions have introduced an automated foreground/background segmentation scheme. Here the indexing of an image is made on the principal shape, with the aid of some heuristics. This is an evident limitation: most images do not have a main shape, and objects are often composed of various parts.

Other researchers, rather than concentrating on a main shape, which is typically assumed located in the central part of the picture, have proposed to index regions in images; so that the focus is not on retrieval of similar images, but of similar regions within an image. Examples of this idea are VISUALSEEK (Smith & Chang, 1996), NETRA (Ma & Manjunath, 1997) and BLOBWORLD (Carson, Thomas, Belongie, Hellerstein, & Malik, 1999). The problem is that although all these systems index regions, they lack of a higher level description of images. Hence, they are not able to describe — and hence query for — more than a single region at a time in an image.

In order to improve retrieval performances, much attention has been paid in recent years to relevance feedback. Relevance feedback is the mechanism, widely used in textual information systems, which allows improving retrieval effectiveness by incorporating the user in the query-retrieval loop. Depending on the initial query the system retrieves a set of documents that the user can mark either as relevant or irrelevant. The system, based on the user preferences, refines the initial query retrieving a new set of documents that should be closer to the user's information need.

This issue is particularly relevant in feature-based approaches, as on one hand, the user lacks of a language to express in a powerful way her information need, but on the other hand, deciding whether an image is relevant or not takes just a glance. Examples of systems using relevance feedback include MARS (Rui, Huang, & Mehrotra, 1997), DRAWSEARCH (Di Sciascio & Mongiello, 1999) and PICHUNTER (Cox et al., 2000).

### 2.2 Approaches Based on Spatial Constraints

This type of approach to the problem of image retrieval concentrates on finding the similarity of images in terms of spatial relations among objects in them. Usually the emphasis is only on relative positions of objects, which are considered as "symbolic images" or icons, identified with a single point in the 2D-space. Information on the content and visual appearance of images are normally neglected.

Chang, Shi, and Yan (1983) present the modeling of this type of images in terms of *2D-strings*, each of the strings accounting for the position of icons along one of the two planar dimensions. In this approach retrieval of images basically reverts to simpler string matching.

Gudivada and Raghavan (1995) consider the objects in a symbolic image associated with vertexes in a weighted graph. Edges — *i.e.*, lines connecting the centroids of a pair of objects — represent the spatial relationships among the objects and are associated with a weight depending on their slope. The symbolic image is represented as an edge list. Given the edge lists of a query and a database image, a similarity function computes the degree of closeness between the two lists as a measure of the matching between the two spatial-graphs. The similarity measure depends on the number of edges and on the comparison between the orientation and slope of edges in the two spatial-graphs. The algorithm is robust with respect to scale and translation variants in the sense that it assigns the highest similarity to an image that is a scale or translation variant of the query image. An extended algorithm includes also rotational variants of the original images.

More recent papers on the topic include those by Gudivada (1998) and by El-Kwae and Kabuka (1999), which basically propose extensions of the strings approach for efficient retrieval of subsets of icons. Gudivada (1998) defines $\theta$R-strings, a logical representation of an image. Such representation also provides a geometry-based approach to iconic indexing based on spatial relationships between the iconic objects in an image individuated by their centroid coordinates. Translation, rotation and scale variant images and the variants generated by an arbitrary composition of these three geometric transformations are considered. The approach does not deal with object shapes, nor with other basic image features, and considers only the sequence of the names of the objects. The concatenation of the objects is based on the euclidean distance of the domain objects in the image starting from a reference point. The similarity between a database and a query image is obtained through a spatial similarity algorithm that measures the degree of similarity between a query and a database image by comparing the similarity between their $\theta$R-strings. The algorithm recognizes rotation, scale and translation variants of the image and also subimages, as subsets of the domain objects. A constraint limiting the practical use of this approach is the assumption that an image can contain at most one instance of each icon or object.

El-Kwae and Kabuka (1999) propose a further extension of the spatial-graph approach, which includes both the topological and directional constraints. The topological extension of the objects can be obviously useful in determining further differences between images that might be considered similar by a directional algorithm that considers only the locations of objects in term of their centroids. The similarity algorithm they propose extends the graph-matching one previously described by Gudivada and Raghavan (1995). The similarity between two images is based on three factors: the number of common objects, the directional and topological spatial constraint between the objects. The similarity measure includes the number of objects, the number of common objects and a function that determines the topological difference between corresponding objects pairs in the query and in the database image. The algorithm retains the properties of the original approach, including its invariance to scaling, rotation and translation and is also able to recognize multiple rotation variants.

### 2.3 Logic-based and Structured Approaches

With reference to previous work on Vision in Artificial Intelligence, the use of structural descriptions of objects for the recognition of their images can be dated back to Minsky's frames, and some work by Brooks (1981). The idea is to associate parts of an object (and generally of a scene) to the regions an image can be segmented into. The hierarchical organization of knowledge to be used in the recognition of an object was first proposed by Marr (1982). Reiter and Mackworth (1989) proposed a formalism to reason about maps as sketched diagrams. In their approach, the possible relative positions of lines are fixed and highly qualitative (touching, intersecting).

Structured descriptions of three-dimensional images are already present in languages for virtual reality like VRML (Hartman & Wernecke, 1996) or hierarchical object modeling. However, the semantics of such languages is operational, and no effort is made to automatically classify objects with respect to the structure of their appearance.

Meghini, Sebastiani, and Straccia (2001) proposed a formalism integrating Description Logics and image and text retrieval, while Haarslev, Lutz, and Möeller (1998) integrate Description Logics with spatial reasoning. Further extensions of the approach are described by Moeller, Neumann, and Wessel (1999). Both proposals build on the clean integration of Description Logics and *concrete domains* of Baader and Hanschke (1991). However, neither of the formalisms can be used to build complex shapes by nesting more simple shapes. Moreover, the proposal by Haarslev et al. (1998) is based on the logic of spatial relations named RCC8, which is enough for specifying meaningful relations in a map, but it is too qualitative to specify the relative sizes and positions of regions in a complex shape.

Also for Hacid and Rigotti (1999) description logics and concrete domains are at the basis of a logical framework for image databases aimed at reasoning on query containment. Unfortunately, the proposed formalism cannot consider geometric transformations neither determine specific arrangements of shapes.

More similar to our approach is the proposal by Ardizzone, Chella, and Gaglio (1997), where parts of a complex shape are described with a description logic. However, the composition of shapes does not consider their positions, hence reasoning cannot take positions into account.

Relative position of parts of a complex shape can be expressed in a constraint relational calculus in the work by Bertino and Catania (1998). However, reasoning about queries (containment and emptiness) is not considered in this approach. Aiello (2001) proposes a multi-modal logic, which provides a formalism for expressing topological properties and for defining a distance measure among patterns.

Spatial relation between parts of medical tomographic images are considered by Tagare, Vos, Jaffe, and Duncan (1995). There, medical images are formed by the intersection of the image plane and an object. As the image plane changes, different parts of the object are considered. Besides, a metric for arrangements is formulated by expressing arrangements in terms of the Voronoi diagram of the parts. The approach is limited to medical image databases and does not provide geometrical constraints.

Compositions of parts of an image are considered in the work by Sanfeliu and Fu (1983) for character recognition. However, in recognizing characters, line compositions are "closed", in the sense that one looks for the specified lines, and no more. Instead in our framework, the shape "F" composed by three lines, is subsumed by the shape "Γ" — something unacceptable in recognizing characters. Apart from the different task, this approach does not make use of an extensional semantics for composite shapes, hence no reasoning is possible.

A logic-based multimedia retrieval system was proposed by Fuhr, Gövert, and Rölleke (1998); the method, based on an object-oriented logic, supports aggregated objects but it is oriented towards a high-level semantic indexing, which neglects low-level features that characterize images and parts of them.

In the field of computation theories of recognition, we mention two approaches that have some resemblance to our own: Biederman's structural decomposition and geometric constraints proposed by Ullman, both described by Edelmann (1999). Unfortunately, neither of them appears suitable for realistic image retrieval: the structural decomposition approach does not consider geometric constraints between shapes, while the approach based on geometric constraints does not consider the possibility of defining structural decomposition of shapes, hence reasoning on them.

Starting with the reasonable assumption that the recognition of an object in a scene can be eased by previous knowledge on the context, in the work by Pirri and Finzi (1999), the recognition task, or the interpretation of an image, takes advantage of the information a cognitive agent has about the environment, and by the representation of these data in a high-level formalism.

## 3. Syntax and Semantics

In this section we present the formalism dealing with the definition of composite shape descriptions, their semantics, and some properties that distinguish our approach from previous ones.

We remark that our formalism deals with image features, like shape, color, texture, but is *independent* of the way features are extracted from actual images. For the interested reader, the algorithms we used to compute image features in our implementation of the formalism are presented in the Appendix.

### 3.1 Syntax

Our main syntactic objects are basic shapes, position of shapes, composite shape descriptions, and transformations. We also take into account the other features that typically determine the visual appearance of an image, namely color and texture.

*Basic shapes* are denoted with the letter $B$, and have an edge contour $e(B)$ characterizing them. We assume that $e(B)$ is described as a single, closed 2D-curve in a space whose origin coincides with the centroid of $B$. Examples of basic shapes can be `circle`, `rectangle`, with the contours $e(\texttt{circle}) = \bigcirc$, $e(\texttt{rectangle}) = \Box$, but also any complete, rough contour — e.g., the one of a ship — is a basic shape. To make our language compositional, we consider only the *external* contour of a region. For example, if a region is contained in another, as in ⊡, the contour of the outer region is just the external rectangle.

The possible transformations are the simple ones that are present in any drawing tool: rotation (around the centroid of the shape), scaling and translation. We globally denote a rotation-translation-scaling transformation as $\tau$. Recall that transformations can be composed in sequences $\tau_1 \circ \ldots \circ \tau_n$, and they form a mathematical group.

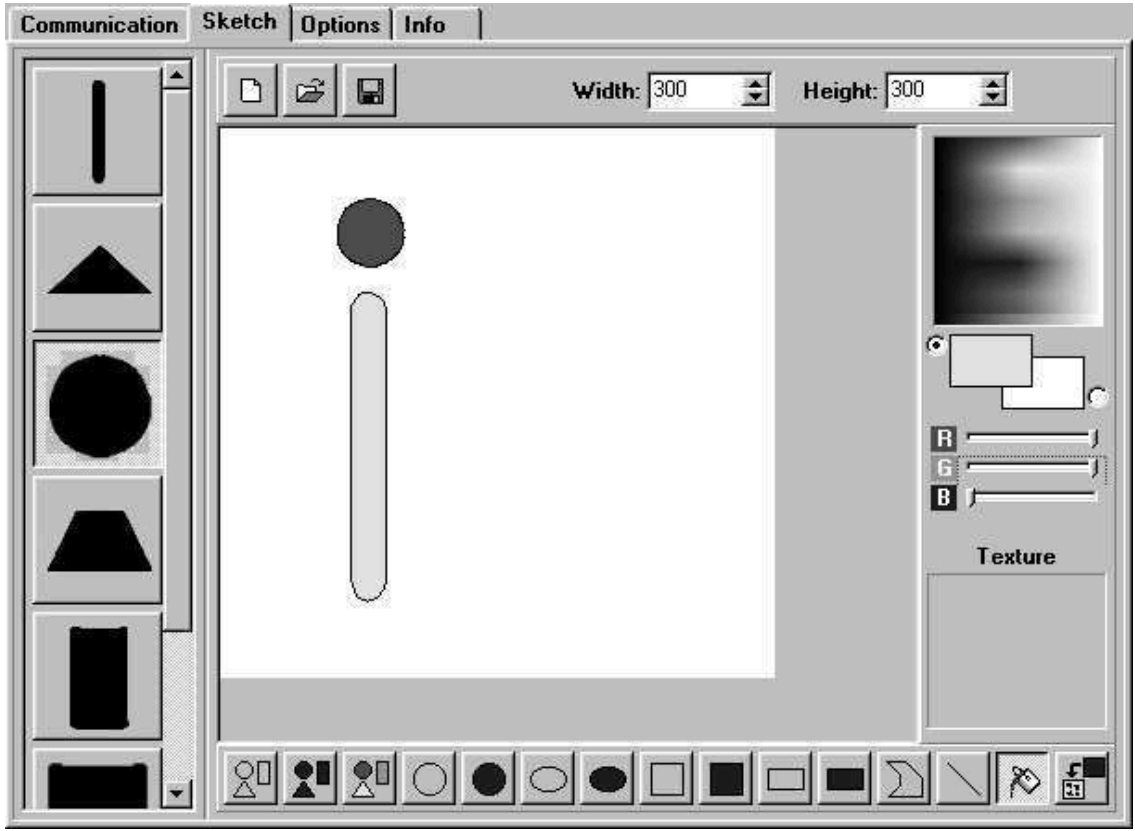

Figure 1: The graphical interface with a query by sketch.

The basic building block of our syntax is a *basic shape component* $\langle c, t, \tau, B \rangle$, which represents a region with color $c$, texture $t$, and edge contour $\tau(e(B))$. With $\tau(e(B))$ we denote the pointwise transformation $\tau$ of the whole contour of $B$. For example, $\tau$ could specify to place the contour $e(B)$ in the upper left corner of the image, scaled by 1/2 and rotated 45 degrees clockwise.

*Composite shape descriptions* are conjunctions of basic shape components — each one with its own color and texture — denoted as

$$C = \langle c_1, t_1, \tau_1, B_1 \rangle \sqcap \cdots \sqcap \langle c_n, t_n, \tau_n, B_n \rangle$$

We do not expect end users of our system to actually define composite shapes with this syntax; this is just the *internal* representation of a composite shape. The system can maintain it while the user draws — with the help of a graphic tool — the complex shape by dragging, rotating and scaling basic shapes chosen either from a palette, or from existing images (see Figure 1).

For example, the composite shape `lighted-candle` could be defined as

$$\texttt{lighted-candle} = \langle c_1, t_1, \tau_1, \texttt{rectangle} \rangle \sqcap \langle c_2, t_2, \tau_2, \texttt{circle} \rangle$$

with $\tau_1, \tau_2$ placing the circle as a flame on top of the candle, and textures and colors defined accordingly to the intuition.

We remark that, to the best of our knowledge, the logic we present is the first one combining shapes and explicit transformations in one language.

In a previous paper (Di Sciascio, Donini, & Mongiello, 2000) we presented a formalism including nested composite shapes, as it is done in hierarchical object modeling (Foley et al., 1996, Ch.7). However, nested composite shapes can always be flattened by composing their transformations. Hence in this paper we focus on two levels: basic shapes and compositions of basic shapes. Also, just to simplify the presentation of the semantics, in the following section we do not present color and texture features, which we take into account from Section 4.2 on.

### 3.2 Semantics

We consider an extensional semantics, in which syntactic expressions are interpreted as subsets of a domain. For our setting, the domain of interpretation is a set of images $\Delta$, and shapes and components are interpreted as subsets of $\Delta$. Hence, also an image database is a domain of interpretation, and a complex shape $C$ is a subset of such a domain — the images to be retrieved from the database when $C$ is viewed as a query.

This approach is quite different from previous logical approaches to image retrieval that view the image database as a set of facts, or logical assertions, *e.g.*, the one based on Description Logics by Meghini et al. (2001). In that setting, image retrieval amounts to logical inference. However, observe that usually a *Domain Closure Assumption* (Reiter, 1980) is made for image databases: there are no regions but the ones which can be seen in the images themselves. This allows one to consider the problem of image retrieval as simple model checking — check if a given structure satisfies a description[2].

Formally, an interpretation is a pair $(\mathcal{I}, \Delta)$, where $\Delta$ is a set of images, and $\mathcal{I}$ is a mapping from shapes and components to subsets of $\Delta$. We identify each image $I$ with the set of regions $\{r_1, \ldots, r_n\}$ it can be segmented into (excluding background, which we discuss at the end of this section). Each region $r$ comes with its own edge contour $e(r)$. An image $I \in \Delta$ belongs to the interpretation of a basic shape component $\langle \tau, B \rangle^{\mathcal{I}}$ if $I$ contains a region whose contour matches $\tau(e(B))$. In formulae,

$$\langle \tau, B \rangle^{\mathcal{I}} = \{I \in \Delta \mid \exists r \in I : e(r) = \tau(e(B))\} \qquad (1)$$

The above definition is only for exact recognition of shape components in images, due to the presence of strict equality in the comparison of contours; but it can be extended to approximate recognition as follows. Recall that the *characteristic function $f_S$* of a set $S$ is a function whose value is either 1 or 0; $f_S(x) = 1$ if $x \in S$, $f_S(x) = 0$ otherwise. We consider now the characteristic function of the set defined in Formula (1). Let $I$ be an image; if $I$ belongs to $\langle \tau, B \rangle^{\mathcal{I}}$, then the characteristic function computed on $I$ has value 1, otherwise it has value 0. To keep the number of symbols low, we use the expression $\langle \tau, B \rangle^{\mathcal{I}}$ also to denote the characteristic function (with an argument ($I$) to distinguish it from the set).

---

2. Obviously, a Domain Closure Assumption on regions is not valid in artificial vision, dealing with two-dimensional images of three-dimensional shapes (and scenes), because solid shapes have surfaces that will be hidden in their images. But this is outside the scope of our retrieval problem.

$$\langle \tau, B \rangle^{\mathcal{I}}(I) = \begin{cases} 1 & \text{if } \exists r \in I : e(r) = \tau(e(B)) \\ 0 & \text{otherwise} \end{cases}$$

Now we reformulate this function in order to make it return a real number in the range $[0, 1]$ — as usual in fuzzy logic (Zadeh, 1965). Let $sim(\cdot, \cdot)$ be a similarity measure from pairs of contours into the range $[0, 1]$ of real numbers (where 1 is perfect matching). We use $sim(\cdot, \cdot)$ instead of equality to compare edge contours. Moreover, the existential quantification can be replaced by a maximum over all possible regions in $I$. Then, the characteristic function for the approximate recognition in an image $I$ of a basic component, is:

$$\langle \tau, B \rangle^{\mathcal{I}}(I) = \max_{r \in I} \{sim(e(r), \tau(e(B)))\}$$

Note that $sim$ depends on translations, rotation and scaling, since we are looking for regions in $I$ whose contour matches $e(B)$, with reference to the position and size specified by $\tau$.

The interpretation of basic shapes, instead, includes a translation-rotation-scaling invariant recognition, which is commonly used in single-shape Image Retrieval. We define the interpretation of a basic shape as

$$B^{\mathcal{I}} = \{I \in \Delta \mid \exists \tau \, \exists r \in I : e(r) = \tau(e(B))\}$$

and its approximate counterpart as the function

$$B^{\mathcal{I}}(I) = \max_{\tau} \max_{r \in I} \{sim(e(r), \tau(e(B)))\}$$

The maximization over all possible transformations $\max_\tau$ can be effectively computed by using a similarity measure $sim_{ss}$ that is invariant with reference to translation-rotation-scaling (see Section 4.2). Similarity of color and texture will be added as a weighted sum in Section 4.2. In this way, a basic shape $B$ can be used as a query to retrieve all images from $\Delta$ which are in $B^{\mathcal{I}}$. Therefore, our approach generalizes the more usual approaches for single-shape retrieval, such as Blobworld (Carson et al., 1999).

Composite shape descriptions are interpreted as sets of images that contain all components of the composite shape. Components can be anywhere in the image, as long as they are in the described arrangement relative to each other. Let $C$ be a composite shape description $\langle \tau_1, B_1 \rangle \sqcap \cdots \sqcap \langle \tau_n, B_n \rangle$. In exact matching, the interpretation is the intersection of the sets interpreting each component of the shape:

$$C^{\mathcal{I}} = \{I \in \Delta \mid \exists \tau : I \in \cap_{i=1}^{n} \langle (\tau \circ \tau_i), B_i \rangle^{\mathcal{I}}\} \qquad (2)$$

Observe that we require all shape components of $C$ to be transformed into image regions using the same transformation $\tau$. This preserves the arrangement of the shape components relative to each other — given by each $\tau_i$ — while allowing $C^{\mathcal{I}}$ to include every image containing a group of regions in the right arrangement, wholly displaced by $\tau$.

To clarify this formula, consider Figure 2: the shape $C$ is composed by two basic shapes $B_1$ and $B_2$, suitably arranged by the transformations $\tau_1$ and $\tau_2$. Suppose now that $\Delta$ contains the image $I$. Then, $I \in C^{\mathcal{I}}$ because there exists the transformation $\tau$, which globally brings $C$ into $I$, that is, $\tau \circ \tau_1$ brings the rectangle $B_1$ into a rectangle recognized in $I$, and $\tau \circ \tau_2$ brings the circle $B_2$ into a circle recognized in $I$, both arranged according to $C$. Note that $I$ could contain also other shapes, not included in $C$.

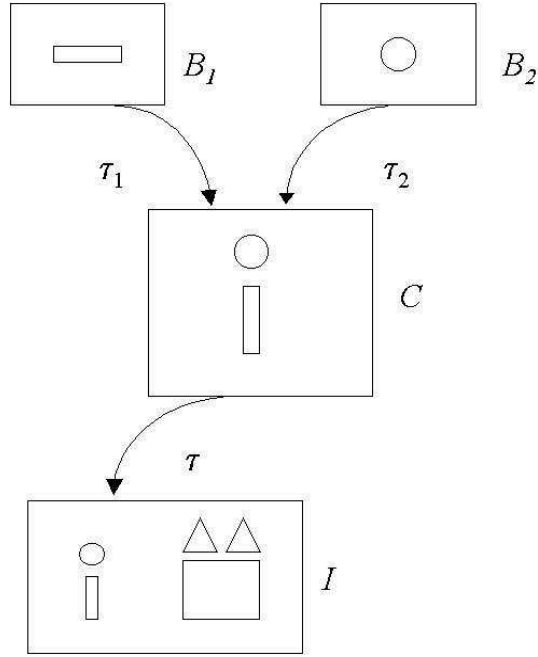

Figure 2: An example of application of Formula (2).

We can now formally define the recognition of a shape in an image.

**Definition 1 (Recognition)** *A shape description $C$ is* recognized *in an image $I$ if for every interpretation $(\mathcal{I}, \Delta)$ such that $I \in \Delta$, it is $I \in C^{\mathcal{I}}$. An interpretation $(\mathcal{I}, \Delta)$ satisfies a composite shape description $C$ if there exists an image $I \in \Delta$ such that $C$ is recognized in $I$. A composite shape description is* satisfiable *if there exists an interpretation satisfying it.*

Observe that shape descriptions could be unsatisfiable: if two components define overlapping regions, no image can be segmented in a way that satisfies both components. Of course, if composite shape descriptions are built using a graphical tool, unsatisfiability can be easily avoided, so we assume that descriptions are always satisfiable. Anyway, unsatisfiable shape descriptions could be easily detected, from their syntactic form, since unsatisfiability can only arise because of overlapping regions (see Proposition 4).

Observe also that our set-based semantics implies the intuitive interpretation of conjunction "⊓" — one could easily prove that ⊓ is commutative and idempotent.

For approximate matching, we modify definition (2), following the fuzzy interpretation of ⊓ as minimum, and existential as maximum:

$$C^{\mathcal{I}}(I) = \max_{\tau} \{\min_{i=1}^{n} \{\langle (\tau \circ \tau_i), B_i \rangle^{\mathcal{I}}(I)\}\} \qquad (3)$$

Observe that our interpretation of composite shape descriptions strictly requires the presence of all components. In fact, the measure by which an image $I$ belongs to the interpreta-tion of a composite shape description $C^\mathcal{I}$ is dominated by the least similar shape component (the one with the minimum similarity). Hence, if a basic shape component is very dissimilar from every region in $I$, this brings near to $0^3$ also the measure of $C^\mathcal{I}(I)$. This is more strict than, *e.g.*, Gudivada & Raghavan's (1995) or El-Kwae & Kabuka's (1999) approaches, in which a non-appearing component can decrease the similarity value of $C^\mathcal{I}(I)$, but $I$ can be still above a threshold.

Although this requirement may seem a strict one, it captures the way details are used to refine a query: the "dominant" shapes are used first, and, if the retrieved set is still too large, the user adds details to restrict the results. In this refinement process, it should not happen that other images that match only some new details, "pop up" *enlarging* the set of results that the user was trying to restrict. We formalize this refinement process through the following definition.

**Proposition 1 (Downward refinement)** *Let $C$ be a composite shape description, and let $D$ be a refinement of $C$, that is $D \doteq C \sqcap \langle \tau', B' \rangle$. For every interpretation $\mathcal{I}$, if shapes are interpreted as in (2), then $D^\mathcal{I} \subseteq C^\mathcal{I}$; if shapes are interpreted as in (3), then for every image $I$ it holds $D^\mathcal{I}(I) \leq C^\mathcal{I}(I)$.*

*Proof.* For (2), the claim follows from the fact that $D^\mathcal{I}$ considers an intersection of the same components as the one of $C^\mathcal{I}$, plus the set $\langle (\tau \circ \tau'), B' \rangle^\mathcal{I}$. For (3), the claim analogously follows from the fact that $D^\mathcal{I}(I)$ computes a minimum over a superset of the values considered for $C^\mathcal{I}(I)$. □

The above property makes our language fully *compositional*. Namely, let $C$ be a composite shape description; we can consider the meaning of $C$ — when used as a query — as the set of images that can be potentially retrieved using $C$. At least, this will be the meaning perceived by an end user of a system. Downward refinement ensures that the meaning of $C$ can be obtained by starting with one component, and then progressively adding other components in any order. We remark that for other frameworks cited above (Gudivada & Raghavan, 1995; El-Kwae & Kabuka, 1999) this property does not hold. We illustrate the problem in Figure 3. Starting with shape description $C$, we may retrieve (among many others) the two images $I_1, I_2$, for which both $C^\mathcal{I}(I_1)$ and $C^\mathcal{I}(I_2)$ are above a threshold $t$, while another image $I_3$ is not in the set because $C^\mathcal{I}(I_3) < t$. In order to be more selective, we try adding details, and we obtain the shape description $D$. Using $D$, we may still retrieve $I_2$, and discard $I_1$. However, $I_3$ now partially matches the new details of $D$. If Downward refinement holds, $D^\mathcal{I}(I_3) \leq C^\mathcal{I}(I_3) < t$, and $I_3$ cannot "pop up". In contrast, if Downward refinement does not hold (as in Gudivada & Raghavan's approach) it can be $D^\mathcal{I}(I_3) > t > C^\mathcal{I}(I_3)$ because matched details in $D$ raise the similarity sum weighted over all components. In this case, the meaning of a sketch cannot be defined in terms of its components.

Downward refinement is a property linking syntax to semantics. Thanks to the extensional semantics, it can be extended to an even more meaningful semantic relation, namely, subsumption. We borrow this definition from Description Logics (Donini, Lenzerini, Nardi, & Schaerf, 1996), and its fuzzy extensions (Yen, 1991; Straccia, 2001).

---

3. Not exactly 0, since every shape matches every other one with a very low similarity measure. Similarity is often computed as the inverse of a distance. Similarity 0 would correspond to infinite distance. Nevertheless, the recognition algorithm can force the similarity to 0 when it is below a threshold.

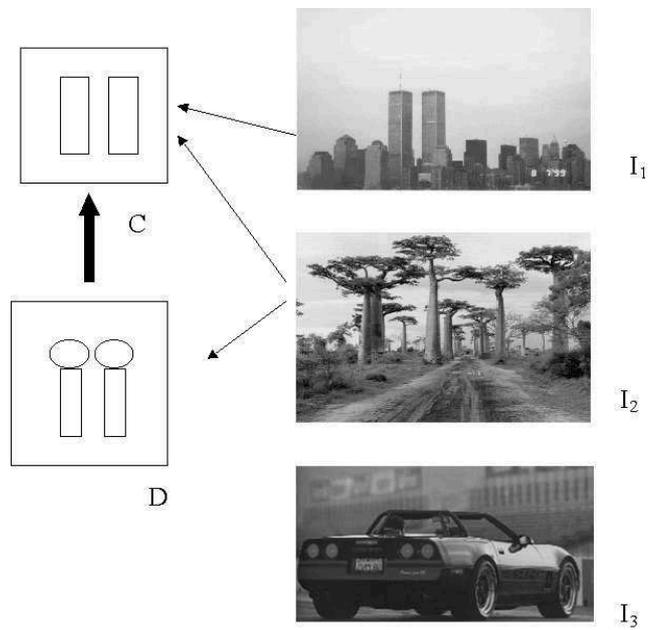

Figure 3: Downward refinement: the thin arrows denote non-zero similarity in approximate recognition. The thick arrow denotes a refinement.

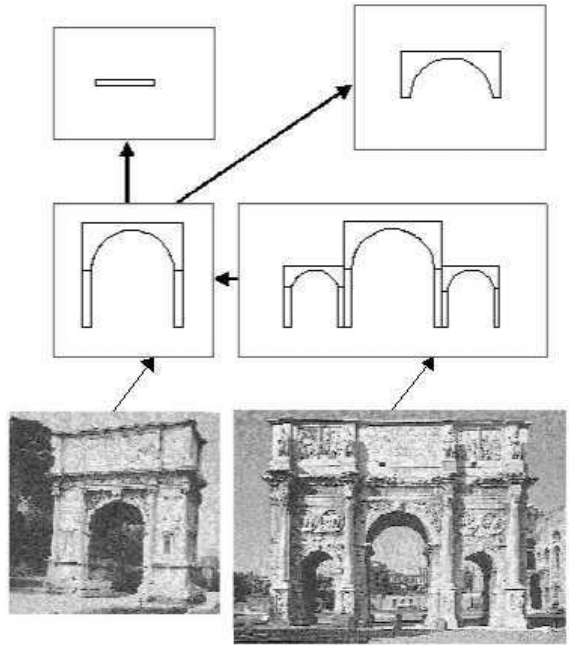

Figure 4: An example of subsumption hierarchy of shapes (thick arrows), and images in which the shapes can be recognized (thin arrows).

**Definition 2 (Subsumption)** *A description $C$ subsumes a description $D$ if for every interpretation $\mathcal{I}$, $D^{\mathcal{I}} \subseteq C^{\mathcal{I}}$. If (3) is used, $C$ subsumes $D$ if for every interpretation $\mathcal{I}$ and image $I \in \Delta$, it is $D^{\mathcal{I}}(I) \leq C^{\mathcal{I}}(I)$.*

Subsumption takes into account the fact that a description might contain a syntactic variant of another, without both the user and the system explicitly knowing this fact. The notion of subsumption extends downward refinement. It enables also a hierarchy of shape descriptions, in which a description $D$ is below another $C$ if $D$ is subsumed by $C$. When $C$ and $D$ are used as queries, the subsumption hierarchy makes easy to detect *query containment*. Containment can be used to speed up retrieval: all images retrieved using $D$ as a query can be immediately retrieved also when $C$ is used as a query, without recomputing similarities. While query containment is important in standard databases (Ullman, 1988), it becomes even more important in an image retrieval setting, since the recognition of specific features in an image can be computationally demanding.

Figure 4 illustrates an example of subsumption hierarchy of basic and composite shapes (thick arrows denote a subsumption between shapes), and two images in which shapes can be recognized (thin arrows).

Although we did not consider a *background*, it could be added to our framework as a special basic component $\langle c, t, \_, \texttt{background} \rangle$ with the property that a region $b$ satisfies the background simply if their colors and textures match, with no check on the edge contours. Also, more than one background could be added; in that case background regions should not overlap, and the matching of background regions should be considered after the regions of all the basic shapes recognized are subtracted to the background regions.

## 4. Reasoning and Retrieval

We envisage several reasoning services that can be carried out in a logic for image retrieval:

1. shape recognition: Given an image $I$ and a shape description $D$, decide if $D$ is recognized in $I$.

2. image retrieval: given a database of images and a shape description $D$, retrieve all images in which $D$ can be recognized.

3. image classification: given an image $I$ and a collection of descriptions $D_1, \ldots, D_n$, find which descriptions can be recognized in $I$. In practice, $I$ is classified by finding the *most specific* descriptions (with reference to subsumption) it satisfies. Observe that classification is a way of "preprocessing" recognition.

4. description subsumption (and classification): given a (new) description $D$ and a collection of descriptions $D_1, \ldots, D_n$, decide whether $D$ subsumes (or is subsumed by) each $D_i$, for $i = 1, \ldots, n$.

While services 1–2 are standard in an image retrieval system, services 3–4 are less obvious, and we briefly discuss them below.

The process of image retrieval is quite expensive, and systems usually perform off-line processing of data, amortizing its cost over several queries to be answered on-line. As an example, all document retrieval systems for the web[4], both for images and text, use spiders to crawl the web and extract some relevant features (*e.g.*, color distributions and textures in images, keywords in texts), that are used to classify documents. Then, the answering process uses such classified, extracted features of documents — and not the original data.

Our system can adapt this setting to composite shapes, too. In our system, a new image inserted in the database is immediately segmented and classified in accordance with the basic shapes that compose it, and the composite descriptions it satisfies (Service 3). Also a query undergoes the same classification, with reference to the queries already answered (Service 4). The more basic shapes are present, the faster will the system answer new queries based on these shapes.

More formally, given a query (shape description) $D$, if there exists a collection of descriptions $D_1, \ldots, D_n$ and all images in the database were already classified with reference to $D_1, \ldots, D_n$, then it may suffice to classify $D$ with reference to $D_1, \ldots, D_n$ to find (most of) the images satisfying $D$. This is the usual way in which classification in Description Logics — which amounts to a semantic indexing — can help query answering (Nebel, 1990).

For example, to answer the query asking for images containing an `arch`, a system may classify `arch` and find that it subsumes `threePortalsGate` (see Figure 4). Then, the system can include in the answer all images in which ancient Roman gates can be recognized, without recomputing whether these images contain an arch or not.

---
[4]. *e.g.*, Altavista, QBIC, NETRA, Blobworld, but also Yahoo (for textual documents).

The problem of computing subsumption between descriptions is reduced to recognition in the next section, and then an algorithm for exact recognition is given. Then, we extend the algorithm to realistic approximate recognition, reconsidering color and texture.

### 4.1 Exact Reasoning on Images and Descriptions

We start with a reformulation of (2), more suited for computational purposes.

**Theorem 2 (Recognition as mapping)** *Let $C = \langle \tau_1, B_1 \rangle \sqcap \cdots \sqcap \langle \tau_n, B_n \rangle$ be a composite shape description, and let $I$ be an image, segmented into regions $\{r_1, \ldots, r_m\}$. Then $C$ is recognized in $I$ iff there exists a transformation $\tau$ and an injective mapping $j : \{1, \ldots, n\} \to \{1, \ldots, m\}$ such that for $i = 1, \ldots, n$ it is*

$$e(r_{j(i)}) = \tau(\tau_i(e(B_i)))$$

*Proof.* From (2), $C$ is recognized in $I$ iff

$$\exists \tau [I \in \bigcap_{i=1}^{n} \langle (\tau \circ \tau_i), B_i \rangle^{\mathcal{I}}] \text{ which is equivalent to } \exists \tau [\bigwedge_{i=1}^{n} I \in \langle (\tau \circ \tau_i), B_i \rangle^{\mathcal{I}}]$$

Expanding $\langle (\tau \circ \tau_i), B_i \rangle^{\mathcal{I}}$ with its definition (1) yields

$$\exists \tau [\bigwedge_{i=1}^{n} \exists r \in I . e(r) = \tau(\tau_i(e(B_i)))]$$

and since regions in $I$ are $\{r_1, \ldots, r_m\}$ this is equivalent to

$$\exists \tau [\bigwedge_{i=1}^{n} \bigvee_{j=1}^{m} e(r_j) = \tau(\tau_i(e(B_i)))]$$

Making explicit the disjunction over $j$ and conjunctions over $i$, we can arrange this conjunctive formula as a matrix:

$$\exists \tau \begin{bmatrix} (e(r_1) = \tau(\tau_1(e(B_1))) & \vee & \cdots & \vee & e(r_m) = \tau(\tau_1(e(B_1)))) & ) & \wedge \\ & \vdots & \vee & \vdots & \vee & \vdots & & \wedge \\ (e(r_1) = \tau(\tau_n(e(B_n))) & \vee & \cdots & \vee & e(r_m) = \tau(\tau_n(e(B_n)))) & ) & \end{bmatrix} \quad (4)$$

Now we note two properties in the above matrix of equalities:

1. For a given transformation, at most one region among $r_1, \ldots, r_m$ can be equal to each component. This means that in each row, at most one disjunct can be true for a given $\tau$.

2. For a given transformation, a region can match at most one component. This means that in each column, at most one equality can be true for a given $\tau$.

We observe that these properties do not imply that regions have all different shapes, since the equality of contours *depends* on any translation, rotation, and scaling. We use equality to represent true overlap, and not just equal shape.

Properties 1–2 imply that the above formula is true iff there is an injective function mapping each component to one region it matches with. To ease the comparison with the formulae above we use the same symbol $j$ as a mapping $j : \{1, \ldots, n\} \to \{1, \ldots, m\}$. Hence, Formula (4) can be rewritten into the claim:

$$\exists \tau [\exists j : \{1..n\} \to \{1..m\} \bigwedge_{i=1}^{n} e(r_{j(i)}) = \tau(\tau_i(e(B_i)))] \qquad (5)$$

$\square$

Hence, even if in the previous section the semantics of a composite shape was derived from the semantics of its components, in computing whether an image contains a composite shape one can focus on groups of regions, one group $r_{j(1)}, \ldots, r_{j(n)}$ for each possible mapping $j$.

Observe that $j$ injective implies $m \geq n$, as one would expect. The above proposition leaves open which one between $\tau$ or $j$ must be chosen first. In fact, in what follows we show that the optimal choice for exact recognition is to mix decisions about $j$ and $\tau$. When approximate recognition will be considered, however, exchanging quantifiers is not harmless. In fact, it can change the order in which approximations are made. We return to this issue in the next section, when we discuss how one can devise algorithms for approximate recognition.

Subsumption in this simple logic for shape descriptions relies on the composition of contours of basic shapes. Intuitively, to actually decide if $D$ is subsumed by $C$, we check if the sketch associated with $D$ — seen as an image — would be retrieved using $C$ as a query. From a logical perspective, the existentially quantified regions in the semantics of shape descriptions (1) are skolemized with their prototypical contours. Formal definitions follow.

**Definition 3 (Prototypical image)** *Let $B$ be a basic shape. Its* prototypical image *is $I(B) = \{e(B)\}$. Let $C = \langle \tau_1, B_1 \rangle \sqcap \cdots \sqcap \langle \tau_n, B_n \rangle$ be a composite shape description. Its* prototypical image *is $I(C) = \{\tau_1(e(B_1)), \ldots, \tau_n(e(B_n))\}$.*

In practice, from a composite shape description one builds its prototypical image just applying the stated transformations to its components (and color/texture fillings, if present). Recall that we envisage this prototypical image to be built directly by the user, with the help of a drawing tool, with basic shapes and colors as palette items. The system will just keep track of the transformations corresponding to the user's actions, and use them in building the (internal) shape descriptions stored with the previous syntax. The feature that makes our proposal different from other query-by-sketch retrieval systems, is precisely that our sketches have also a logical meaning. So, properties about description/sketches can be proved, containment between query sketches can be stated in a formal way, and algorithms for containment checking can be proved correct with reference to the semantics.

Prototypical images have some important properties. The first is that they satisfy (in the sense of Definition 1) the shape description they exemplify — as intuition would suggest.

**Proposition 3** *For every composite shape description D, if D is satisfiable then the interpretation $\langle \mathcal{I}, \{I(D)\} \rangle$ satisfies D.*

*Proof.* From Theorem 2, using an identical transformation $\tau$ and the identity mapping for $j$. □

A shape description $D$ is satisfiable if there are no overlapping regions in $I(D)$. Since this is obvious when $D$ is specified by a drawing tool, we just give the following proposition for sake of completeness.

**Proposition 4** *A shape description D is satisfiable iff its prototypical image I(D) contains no overlapping regions.*

We now turn to subsumption. Observe that if $B_1$ and $B_2$ are basic shapes, either they are equivalent (each one subsumes the other) or neither of the two subsumes the other. If we adopt for the segmented regions an invariant representation, (e.g. Fourier transforms of the contour) deciding equivalence between basic shapes, or recognizing whether a basic shape appears in an image, is just a call to an algorithm computing the similarity between shapes. This is what usual image recognizers do — allowing for some tolerance in the matching of the shapes. Therefore, our framework extends the retrieval of shapes made of a single component, for which effective systems are already available.

We now consider composite shape descriptions, and prove the main property of prototypical images, namely, the fact that subsumption between shape descriptions can be decided by checking if the subsumer can be recognized in the sketch of the subsumee.

**Theorem 5** *A composite shape description C subsumes a description D if and only if C is recognized in the prototypical image I(D).*

*Proof.* Let $C = \langle \tau_1, B_1 \rangle \sqcap \cdots \sqcap \langle \tau_n, B_n \rangle$, and let $D = \langle \sigma_1, A_1 \rangle \sqcap \cdots \sqcap \langle \sigma_m, A_m \rangle$. Recall that $I(D)$ is defined by $I(D) = \{\sigma_1(e(A_1)), \ldots, \sigma_m(e(A_m))\}$. To ease the reading, we sketch the idea of the proof in Figure 5.

**If.** Suppose $C$ is recognized in $I(D)$, that is, $I(D) \in C^{\mathcal{I}}$ for every interpretation $(\mathcal{I}, \Delta)$ such that $I(D) \in \Delta$. Then, from Theorem 2 there exists a transformation $\hat{\tau}$ and a suitable injective function $j$ from $\{1, \ldots, n\}$ into $\{1, \ldots, m\}$ such that

$$e(r_{j(k)}) = \hat{\tau} \circ \tau_k(e(B_k)) \qquad \text{for } k = 1, \ldots, n$$

Since $I(D)$ is the prototypical image of $D$, we can substitute each region with the basic shape of $D$ it comes from:

$$\sigma_{j(k)}(e(A_{j(k)})) = \hat{\tau} \circ \tau_k(e(B_k)) \qquad \text{for } k = 1, \ldots, n \qquad (6)$$

Now suppose that $D$ is recognized in an image $J = \{s_1, \ldots, s_p\}$, with $J \in \Delta$. We prove that also $C$ is recognized in $J$. In fact, if $D$ is recognized in $J$ then there exists a transformation $\hat{\sigma}$ and another injective mapping $q$ from $\{1, \ldots, m\}$ into $\{1, \ldots, p\}$ selecting from $J$ regions $\{s_{q(1)}, \ldots, s_{q(m)}\}$ such that

$$e(s_{q(h)}) = \hat{\sigma} \circ \sigma_h(e(A_h)) \qquad \text{for } h = 1, \ldots, m \qquad (7)$$

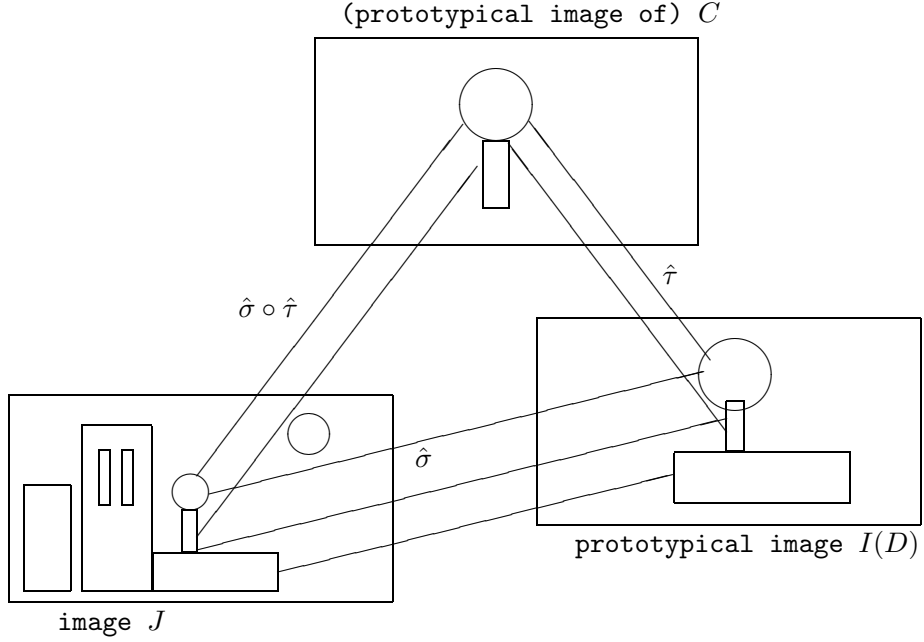

Figure 5: A sketch of the If-proof of Theorem 5

Now composing $q$ and $j$ — that is, selecting the regions of $J$ satisfying those components of $D$ which are used to recognize $C$ — one obtains

$$e(s_{q(j(k))}) = \hat{\sigma} \circ \sigma_{j(k)}(e(A_{j(k)})) \qquad \text{for } k = 1, \ldots, n \tag{8}$$

Then, substituting equals for equals from (6), one finally gets

$$e(s_{q(j(k))}) = \hat{\sigma} \circ \hat{\tau} \circ \tau_k(e(B_k)) \qquad \text{for } k = 1, \ldots, n$$

which proves that $C$ too is recognized in $J$, using $\hat{\sigma} \circ \hat{\tau}$ as transformation of its components, and $q(j(\cdot))$ as injective mapping from $\{1, \ldots, n\}$ into $\{1, \ldots, p\}$. Since $J$ is a generic image, it follows that $D^{\mathcal{I}} \subseteq C^{\mathcal{I}}$. Since $(\mathcal{I}, \Delta)$ is generic too, $C$ subsumes $D$.

**Only if.** The reverse direction is easier: suppose $C$ subsumes $D$. By definition, this amounts to $D^{\mathcal{I}} \subseteq C^{\mathcal{I}}$ for every collection of images $\mathcal{I}$. For every $\mathcal{I}$ that contains $I(D)$, then $I(D) \in D^{\mathcal{I}}$ for Proposition 3. Therefore, $I(D) \in C^{\mathcal{I}}$, that is, $C$ is recognized in $I(D)$. □

This property allows us to compute subsumption as recognition, so we concentrate on complex shape recognition, using Theorem 2. Our concern is how to decide whether there exists a transformation $\tau$ and a matching $j$ having the properties stated in Theorem 2. It turns out that for exact recognition, a quadratic upper bound can be attained for the possible transformations to try.

**Theorem 6** *Let $C = \langle \tau_1, B_1 \rangle \sqcap \cdots \sqcap \langle \tau_n, B_n \rangle$ be a composite shape description, and let $I$ be an image, segmented into regions $\{r_1, \ldots, r_m\}$. Then, there are at most $m(m-1)$ exact matches between the $n$ basic shapes and the $m$ regions. Moreover, each possible match can be verified by checking the matching of $n$ pairs of contours.*

*Proof.* A transformation $\tau$ matching exactly basic components to regions is also an exact match for their centroids. Hence we concentrate on centroids. Each correspondence between a centroid of a basic component and a centroid of a region yields two constraints for $\tau$. Now $\tau$ is a rigid motion with scaling, hence it has four degrees of freedom (two degrees for translations, one for rotation, and one for uniform scaling). Hence, if an exact match $\tau$ exists between the centroids of the basic components and the centroids of some of the regions, then $\tau$ is completely determined by the transformation of any two centroids of the basic shapes into two centroids of the regions.

Fixing any pair of basic components $B_1, B_2$, let $\mathbf{p}_1, \mathbf{p}_2$ denote their centroids. Also, let $r_{j(1)}, r_{j(2)}$ be the regions that correspond to $B_1, B_2$, and let $\mathbf{v}_{j(1)}, \mathbf{v}_{j(2)}$, denote their centroids. There is only one transformation $\tau$ solving the point equations (each one mapping a point into another)

$$\begin{cases} \tau(\tau_1(\mathbf{p}_1)) &= \mathbf{v}_{j(1)} \\ \tau(\tau_2(\mathbf{p}_2)) &= \mathbf{v}_{j(2)} \end{cases}$$

Hence, there are only $m(m-1)$ such transformations. For the second claim, once a $\tau$ matching the centroids is found, one checks that the *edge contours* of basic components and regions coincide, i.e., that $\tau(\tau_1(e(B_1))) = e(r_{j(1)})$, $\tau(\tau_2(e(B_2))) = e(r_{j(2)})$, and for $k = 3, \ldots, n$ that $\tau(\tau_k(e(B_k)))$ coincides with the contour of some region $e(r_{j(k)})$.

Recalling Formula (5) in the proof of Theorem 2, this means that we can eliminate the outer quantifier in (5) using a computed $\tau$, and conclude that $C$ is recognized in $I$ iff:

$$\exists j : \{1..n\} \to \{1..m\} \bigwedge_{i=1}^{n} e(r_{j(i)}) = \tau(\tau_i(e(B_i)))$$

□

Observe that, to prune the above search, once a $\tau$ has been found as above, one can check for $k = 3, \ldots, n$ that $\tau(\tau_k(centr(B_k)))$ coincides with a centroid of some region $r_j$, before checking contours.

Based on Theorem 6, we can devise the following algorithm:

**Algorithm** Recognize $(C,I)$;
**input** a composite shape description $C = \langle \tau_1, B_1 \rangle \sqcap \cdots \sqcap \langle \tau_n, B_n \rangle$, and
an image $I$, segmented into regions $r_1, \ldots, r_m$
**output** True if $C$ is recognized in $I$, False otherwise
**begin**
(1) compute the centroids $\mathbf{v}_1, \ldots, \mathbf{v}_m$ of $r_1, \ldots, r_m$
(2) compute the centroids $\mathbf{p}_1, \ldots, \mathbf{p}_n$ of the components of $C$
(3) **for** $i, h \in \{1, \ldots, m\}$ with $i < h$ **do**
   compute the transformation $\tau$ such that $\tau(\mathbf{p}_1) = \mathbf{v}_i$ and $\tau(\mathbf{p}_2) = \mathbf{v}_h$;

STRUCTURED KNOWLEDGE REPRESENTATION FOR IMAGE RETRIEVAL

```
        if for every k ∈ {1, . . . , n}
            τ(τ_k(e(B_k))) coincides (for some j) with a region r_j in I
        then return True
    endfor
    return False
end
```

The correctness of Recognize $(C,I)$ follows directly from Theorems 2 and 6. Regarding the time complexity, step (1) requires to compute centroids of segmented regions. Several methods for computing centroids are well known in the literature (Jahne, Haubecker, & Geibler, 1999). Hence, we abstract from this detail, and assume there exists a function $f(N_h, N_v)$ that bounds the complexity of computing one centroid, where $N_h, N_v$ are the horizontal and vertical dimensions of $I$ (number of pixels). We report in the Appendix how we compute centroids, and concentrate on the complexity in terms of $n$, $m$, and $f(N_h, N_v)$.

**Theorem 7** *Let $C = \langle \tau_1, B_1 \rangle \sqcap \cdots \sqcap \langle \tau_n, B_n \rangle$ be a composite shape description, and let $I$ be an image with $N_h \times N_v$ pixels, segmented into regions $\{r_1, \ldots, r_m\}$. Moreover, let $f(N_h, N_v)$ be a function bounding the complexity of computing the centroid of one region. Then $C$ can be recognized in $I$ in time $O(m \cdot f(N_h, N_v) + n + m^2 \cdot n \cdot N_h \cdot N_v)$.*

*Proof.* From the assumptions, Step (1) can be performed in time $O(m \cdot f(N_h, N_v))$. Instead, Step (2) can be accomplished by extracting the $n$ translation vectors from the transformations $\tau_1, \ldots, \tau_n$ of the components of $C$. Therefore, it requires $O(n)$ time. Finally, the innermost check in Step (3) — checking whether a transformed basic shape and a region coincide — can be performed in $O(N_h \cdot N_v)$, using a suitable marking of pixels in $I$ with the region they belong to. Hence, we obtain the claim. □

Since subsumption between two shape descriptions $C$ and $D$ can be reduced to recognizing $C$ in $I(D)$, the same upper bound holds for checking subsumption between composite shape descriptions, with the simplification that also Step (1) can be accomplished without any further feature-level image processing.

### 4.2 Approximate Recognition

The algorithm proposed in the previous section assumes an exact recognition. Since the target of retrieval are real images, approximate recognition is needed. We start by reconsidering the proof of Theorem 2, and in particular the matrix of equalities (4). Using the semantics for approximate recognition (3), the expanded formula for evaluating $C^{\mathcal{I}}(I)$ becomes now the following:

$$\max_{\tau} \min \left\{ \begin{array}{cccc} \max\{sim(e(r_1), \tau(\tau_1(e(B_1)))), & \ldots, & sim(e(r_m), \tau(\tau_1(e(B_1)))) & \} \\ \vdots & \vdots & \vdots & \\ \max\{sim(e(r_1), \tau(\tau_n(e(B_n))), & \ldots, & sim(e(r_m), \tau(\tau_n(e(B_n)))) & \} \end{array} \right\}$$

Now Properties 1–2 stated for exact recognition can be reformulated as hypotheses about $sim$, as follows.

1. For a given transformation, we assume that at most one region among $r_1,\ldots,r_m$ is maximally similar to each component. This assumption can be justified by supposing its negation: if there are two regions both maximally similar to a component, then this maximal value should be a very low one, lowering the overall value because of the external minimization. This means that in maximizing each row, we can assume that the maximal value is given by one index among $1,\ldots,m$.

2. For a given transformation, we assume that a region can yield a maximal similarity for at most one component. Again, the rationale of this assumption is that when a region yields a maximal similarity with two components in two different rows, this value can be only a low one, which propagates along the overall minimum. This means that in minimizing the maxima from all rows, we can consider a different region in each row.

We remark that also in the approximate case these assumptions do not imply that regions have all different shapes, since $sim$ is a similarity measure which is 1 only for true overlap, not just for equal shapes with different pose. The assumptions just state that $sim$ should be a function "near" to plain equality.

The above assumptions imply that we can focus on injective mappings from $\{1..n\}$ into $\{1..m\}$ also for the approximate recognition, yielding the formula

$$\max_{\tau} \max_{j:\{1..n\}\to\{1..m\}} \min_{i=1}^{n}\{sim(e(r_{j(i)}), \tau(\tau_i(e(B_i))))\}$$

The choices of $\tau$ and $j$ for the two maxima are independent, hence we can consider groups of regions first:

$$\max_{j:\{1..n\}\to\{1..m\}} \max_{\tau} \min_{i=1}^{n}\{sim(e(r_{j(i)}), \tau(\tau_i(e(B_i))))\} \qquad (9)$$

Differently from the exact recognition, the choice of an injective mapping $j$ does not directly lead to a transformation $\tau$, since now $\tau$ depends on how the similarity of transformed shapes is computed, that is, the choice of $\tau$ depends on $sim$.

In giving a definition of $sim$, we reconsider the other image features (color, texture) that were skipped in the theoretical part to ease the presentation of semantics. This will introduce weighted sums in the similarity measure, where weights are set by the user according to the importance of the features in the recognition.

Let $sim(r, \langle c,t,\tau,B\rangle)$ be a similarity measure that takes a region $r$ (with its color $c(r)$ and texture $t(r)$) and a component $\langle c,t,\tau,B\rangle$ into the range $[0,1]$ of real numbers (where 1 is perfect matching). We note that color and texture similarities do not depend on transformations, hence their introduction does not change Assumptions 1–2 above. Accordingly, Formula (9) becomes

$$\max_{j:\{1..n\}\to\{1..m\}} \max_{\tau} \min_{i=1}^{n}\{sim(r_{j(i)}, \langle c,t,(\tau\circ\tau_i), B_i\rangle)\} \qquad (10)$$

This formula suggests that from all the groups of regions in an image that might resemble the components, we should select the groups that present the higher similarity. In artificially constructed examples in which all shapes in $I$ and $C$ resemble each other, this may generate an exponential number of groups to be tested. However, we can assume that in realistic images the similarity between shapes is selective enough to yield only a very small number of possible groups to try. We recall that in Gudivada's approach (Gudivada, 1998) an even stricter assumption is made, namely, each basic component in $C$ does not appear twice, and each region in $I$ matches at most one component in $C$. Hence our approach extends Gudivada's one, also for this aspect — besides the fact that we consider shape, scale, rotation, color and texture of each component.

In spite of the assumptions made, finding an algorithm for computing the "best" $\tau$ in Formula (10) proved for us a difficult task. The problem is that there is a continuous spectrum of $\tau$ to be searched, and that the best $\tau$ may not be unique. We observed that when only single points are to be matched — instead of regions and components — our problem simplifies to Point Pattern Matching in Computational Geometry. However, even recent results in that research area are not complete, and cannot be directly applied to our problem. Cardoze and Schulman (1998) solve the nearly-exact point matching with efficient randomized methods, but without scaling. They also observe that best match is a more difficult problem than nearly-exact match. Also Chew, Goodrich, Huttenlocher, Kedem, Kleinberg, and Kravets (1997) propose a method for best match of shapes, but they analyze only rigid motions without scaling.

Therefore, we adopt some heuristics to evaluate the above formula. First of all, we decompose $sim(r, \langle c, t, \tau, B \rangle)$ as a sum of six weighted contributions.

Three contributions are independent of the pose: color, texture and shape. The values of color and texture similarity are denoted by $sim_{color}(c(r), c)$ and $sim_{texture}(t(r), t)$, respectively. Similarity of the shapes (rotation-translation-scale invariant) is denoted by $sim_{shape}(e(r), e(B))$. For each feature, and each pair (region, component) we compute a similarity measure as explained in the Appendix. Then, we assign to all similarities of a feature — say, color — the worst similarity in the group. This yields a pessimistic estimate of Formula (10); however, for such estimate the Downward Refinement property holds (see next Theorem 8).

The other three contributions depend on the pose, and try to evaluate how the pose of each region in the selected group is similar to the pose specified by the corresponding component in the sketch. In particular, $sim_{scale}(e(r), \tau(e(B)))$ represents how similar in scale are the region and the transformed component, while $sim_{rotation}(e(r), \tau(e(B)))$ denotes how $e(r)$ and $\tau(e(B))$ are similarly (or not) rotated with reference to the arrangement of the other components. Finally, $sim_{spatial}(e(r), \tau(e(B)))$ denotes a measure of how coincident are the centroids of the region and the transformed component.

In summary, we get the following form for the overall similarity between a region and a component:

$$\begin{aligned}
sim(r, \langle c, t, \tau, B \rangle) &= sim_{spatial}(e(r), \tau(e(B))) \cdot \alpha + \\
&\quad sim_{shape}(e(r), e(B)) \cdot \beta + \\
&\quad sim_{color}(c(r), c) \cdot \gamma + \\
&\quad sim_{rotation}(e(r), \tau(e(B))) \cdot \delta + \\
&\quad sim_{scale}(e(r), \tau(e(B))) \cdot \eta + \\
&\quad sim_{texture}(t(r), t) \cdot \epsilon
\end{aligned}$$

where coefficients $\alpha, \beta, \gamma, \delta, \eta, \epsilon$ weight the relevance each feature has in the overall similarity computation. Obviously, we impose $\alpha + \beta + \gamma + \delta + \eta + \epsilon = 1$, and all coefficients are greater or equal to 0. The actual values given to these coefficients in the implemented system are reported in Table 2 in Section 6.

Because of the difficulties in computing the best $\tau$, we do not compute a maximum over all possible $\tau$'s. Instead, we evaluate whether there *can be* a rigid transformation with scaling from $\tau_1(e(B_1)), \ldots, \tau_n(e(B_n))$ into $r_{j(1)}, \ldots, r_{j(n)}$, through similarities $sim_{spatial}, sim_{scale}$, and $sim_{rotation}$. There is a transformation iff all these similarities are 1. If not, the lower the similarities are, the less "rigid" the transformation should be to match components and regions. Hence, instead of Formula (10) we evaluate the following simpler formula:

$$\max_{j:\{1..n\}\to\{1..m\}} \min_{i=1}^{n}\{sim(r_{j(i)}, \langle c, t, \tau_i, B_i\rangle)\} \qquad (11)$$

interpreting pose similarities in a different way. We now describe in detail how we estimate pose similarities.

Let $C = \langle c_1, t_1, \tau_1, B_1\rangle) \sqcap \cdots \sqcap \langle c_n, t_n, \tau_n, B_n\rangle)$, and let $j$ be an injective function from $\{1..n\}$ into $\{1..m\}$, that matches components with regions $\{r_{j(1)}, \ldots, r_{j(n)}\}$ respectively.

### 4.2.1 Spatial Similarity

For a given component — say, component 1 — we compute all angles under which the other components are seen from 1. Formally, let $\widehat{\alpha_{i1h}}$ be the counter-clockwise-oriented angle with vertex in the centroid of component 1, and formed by the lines linking this centroid with the centroids of component $i$ and $h$. There are $n(n-1)/2$ such angles.

Then, we compute the correspondent angles for region $r_{j(1)}$, namely, angles $\widehat{\beta_{j(i)j(1)j(h)}}$ with vertex in the centroid of $r_{j(1)}$, formed by the lines linking this centroid with the centroids of regions $r_{j(i)}$ and $r_{j(h)}$ respectively. A pictorial representation of the angles is given in Figure 6.

Then we let the difference $\Delta_{spatial}(e(r_{j(1)}), \tau_1(e(B_1)))$ be the maximal absolute difference between correspondent angles:

$$\Delta_{spatial}(e(r_{j(1)}), \tau_1(e(B_1))) = \max_{i,h=2,\ldots,n, i\neq h}\{|\widehat{\alpha_{i1h}} - \widehat{\beta_{j(i)j(1)j(h)}}|\}$$

We compute an analogous measure for components $2, \ldots, n$, and then we select the maximum of such differences:

$$\Delta_{spatial}[j] = \max_{i=1}^{n}\{\Delta_{spatial}(e(r_{j(i)}), \tau_i(e(B_i)))\} \qquad (12)$$

where the argument $j$ highlights the fact that this measure depends on the mapping $j$. Finally, we transform this maximal difference — for which perfect matching yields 0 — into a minimal similarity — perfect matching yields 1 — with the help of the function $\Phi$ described in the Appendix. This minimal similarity is then assigned to every $sim_{spatial}(e(r_{j(i)}), \tau_i(e(B_i)))$, for $i = 1, \ldots, n$.

Intuitively, our estimate measures the difference in the arrangement of centroids between the composite shape and the group of regions. If there exists a transformation bringing components into regions exactly, every difference is 0, and so $sim_{spatial}$ raises to 1 for every component. The more an arrangement is scattered with reference to the other arrangement, the higher its maximum difference. The reason why we use the maximum of all differences as similarity for each pair component-region will be clear when we prove later that this measure obeys Downward Refinement property.

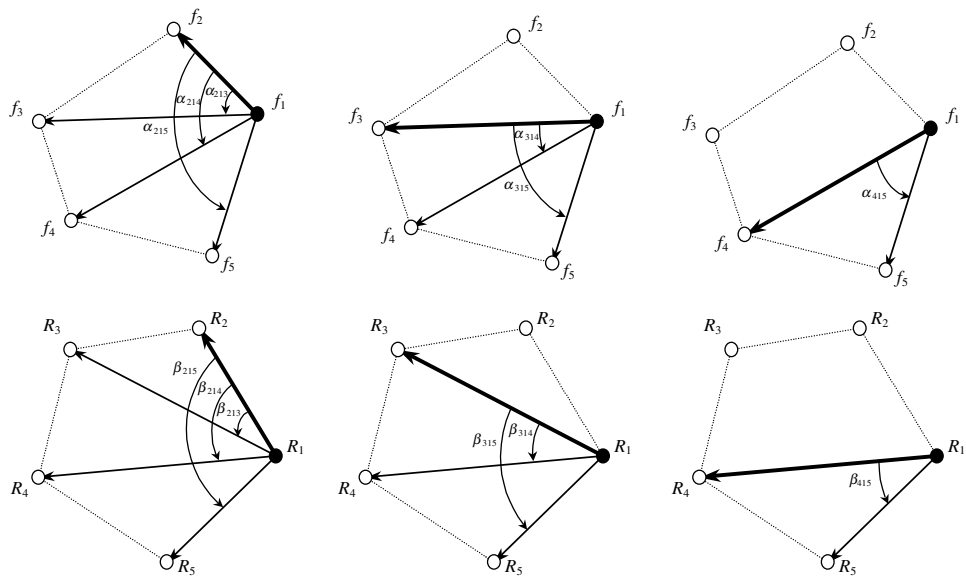

Figure 6: Representation of angles used for computing spatial similarity of component 1 and region $r_{j(1)}$.

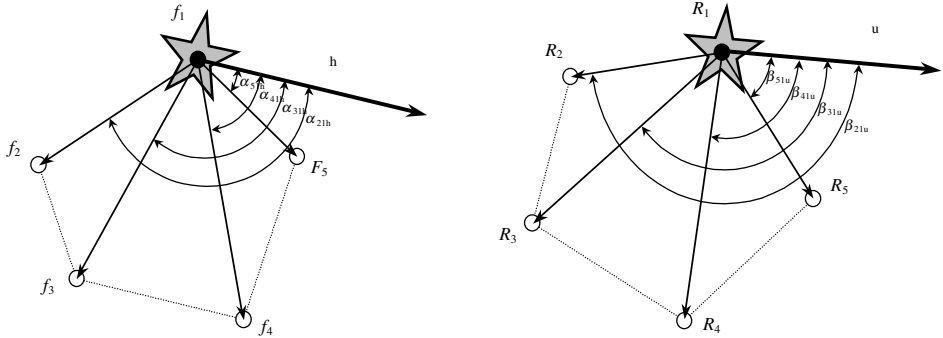

Figure 7: Representation of angles used for computing rotation similarity of component 1 and region $r_{j(1)}$.

### 4.2.2 Rotation Similarity

For every basic shape one can imagine a unit vector with origin in its centroid and oriented horizontally on the right (as seen on the palette). When the shape is used as a component — say, component 1 — also this vector is rotated according to $\tau_1$. Let $\vec{h}$ denote such a rotated vector. For $i = 2, \ldots, n$ let $\gamma_{\widehat{i1\vec{h}}}$ the counter-clockwise-oriented angle with vertex in the centroid of component 1, and formed by $\vec{h}$ and the line linking the centroid of component 1 with the centroid of component $i$.

For region $r_{j(1)}$, the analogous $\vec{u}$ of $\vec{h}$ can be constructed by finding the rotation phase for which cross-correlation attains a maximum value (see Appendix). Then, for $i = 2, \ldots, n$ let $\delta_{\widehat{j(i)j(1)\vec{u}}}$ be the angles with vertex in the centroid of $r_{j(1)}$, and formed by $\vec{u}$ and the line linking the centroid of $r_{j(1)}$ with the centroid of $r_{j(i)}$. Figure 7 clarifies the angles we are computing.

Then we let the difference $\Delta_{rotation}(e(r_{j(1)}), \tau_1(e(B_1)))$ be the maximal absolute difference between correspondent angles:

$$\Delta_{rotation}(e(r_{j(1)}), \tau_1(e(B_1))) = \max_{i=2,\ldots,n} \{|\gamma_{\widehat{i1\vec{h}}} - \delta_{\widehat{j(i)j(1)\vec{u}}}|\}$$

If there is more than one orientation of $r_{j(1)}$ for which cross-correlation yields a maximum — e.g., a square has four such orientations — then we compute the above maximal difference for all such orientations, and take the best difference (the minimal one).

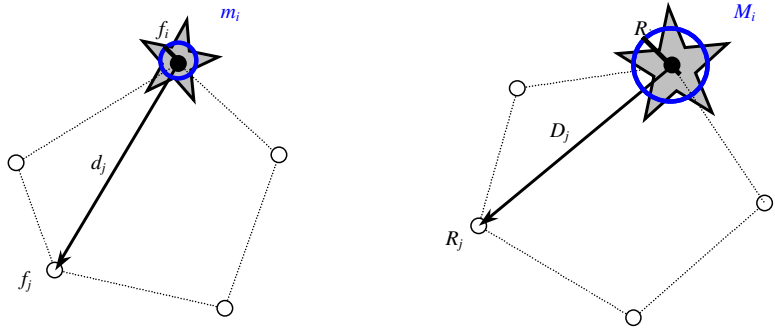

Figure 8: Sizes and distances for scale similarity computation of component 1 and region $r_{j(1)}$.

We repeat the process for components 2 to $n$, and we select the maximum of such differences:

$$\Delta_{rotation}[j] = \max_{i=1}^{n}\{\Delta_{rotation}(e(r_{j(i)}), \tau_i(e(B_i)))\} \tag{13}$$

Finally, as for spatial similarity, we transform $\Delta_{rotation}[j]$ into a minimal similarity with the help of $\Phi$. This minimal similarity is then assigned to every $sim_{rotation}(e(r_{j(i)}), \tau_i(e(B_i)))$, for $i = 1, \ldots, n$.

Observe that also these differences drop to 0 when there is a perfect match, hence the similarity raises to 1. The more a region has to be rotated with reference to the other regions to match a component, the higher the rotational differences. Again, the fact that we use the worst difference to compute all rotational similarities will be exploited in the proof of Downward Refinement.

4.2.3 SCALE SIMILARITY

We concentrate again on component 1 to ease the presentation. Let $m_1$ be the *size* of component 1, computed as the mean distance between its centroid and points on the contour. Moreover, for $i = 2, \ldots, n$, let $d_{1i}$ be the distance between the centroid of component 1 and the centroid of component $i$. In the image, let $M_{j(1)}$ be the size of region $r_{j(i)}$, and let $D_{j(1)j(i)}$ be the distance between centroids of regions $j(1)$ and $j(i)$. Figure 8 pictures the quantities we are computing.

We define the difference in scale between $e(r_{j(1)})$ and $\tau_1(e(B_1))$ as:

$$\Delta_{scale}(e(r_{j(1)}), \tau_1(e(B_1))) = \max_{i=2,\ldots,n}\left\{\left|1 - \frac{\min\{M_{j(1)}/D_{j(1)j(i)}, m_1/d_{1i}\}}{\max\{M_{j(1)}/D_{j(1)j(i)}, m_1/d_{1i}\}}\right|\right\}$$

We repeat the process for components 2 to $n$, and we select the maximum of such differences:

$$\Delta_{scale}[j] = \max_{i=1}^{n}\{\Delta_{scale}(e(r_{j(i)}), \tau_i(e(B_i)))\} \tag{14}$$

Finally, as for the other similarities, we transform $\Delta_{scale}[j]$ into a minimal similarity with the help of $\Phi$. This minimal similarity is then assigned to every $sim_{scale}(e(r_{j(i)}), \tau_i(e(B_i)))$, for $i = 1, \ldots, n$.

### 4.2.4 Discussion of Pose Similarities

Using the same worst difference in evaluating pose similarities of all components may appear a somewhat drastic choice. However, we were guided in this choice by the goal of preserving the Downward Refinement property, even if we had to abandon the exact recognition of the previous section.

**Theorem 8** *Let $C$ be a composite shape description, and let $D$ be a refinement of $C$, that is, $D \doteq C \sqcap \langle c', t', \tau', B' \rangle$. For every image $I$, segmented into regions $r_1, \ldots, r_m$, if $C^{\mathcal{I}}(I)$ and $D^{\mathcal{I}}(I)$ are computed as in (11) using similarities defined above, then it holds $D^{\mathcal{I}}(I) \leq C^{\mathcal{I}}(I)$.*

*Proof.* Every injective function $j$ used to map components of $C$ into $I$ can be extended to a function $j'$ by letting $j'(n+1) \in \{1, \ldots, m\}$ be a suitable region index not in the range of $j$. Since $D^{\mathcal{I}}(I)$ is computed over such extended mappings, it is sufficient to show that values computed in Formula (11) do not increase with reference to the values computed for $C$.

Let $j_1$ be the mapping for which the maximum value $C^{\mathcal{I}}(I)$ is reached. Every extension $j'_1$ of $j_1$ leads to a minimum value $\min_{i=1}^{n+1}$ in Formula (11) which is lower than $C^{\mathcal{I}}(I)$. In fact, all pose differences (12), (13), (14), are computed as maximums over a strictly greater set of values, hence the pose similarities have either the same value, or a lower one. Regarding color, texture, and shape similarities, adding another component can only worsen the values for components of $C$, since we assign to all components the worst similarity in the group.

Now consider another injective mapping $j_2$ that yields a non-maximum value $v_2 < C^{\mathcal{I}}(I)$ in Formula (11). Using the above argument about pose differences (12), (13), (14), every extension $j'_2$ leads to a minimum value $v'_2 \leq v_2$. Since $v_2 < C^{\mathcal{I}}(I)$, also every extension of every mapping $j$ different from $j_1$ yields a value which is less than $C^{\mathcal{I}}(I)$. This completes the proof. $\square$

## 5. A Prototype System

In order to substantiate our ideas we have developed a prototype system, written in C++. The system is a client-server application working in a MS-Windows environment.

The client side avails of a graphical user interface that allows one to carry out all the operations necessary to query the knowledge base, including a canvas for query by sketch composition using basic shapes and a module for query by example using new or existing images as queries. The client also allows a user to insert new shape descriptions and images in the knowledge base. The client has the logical structure shown in Figure 9. It is made up of three main modules: sketch, communication and configuration.

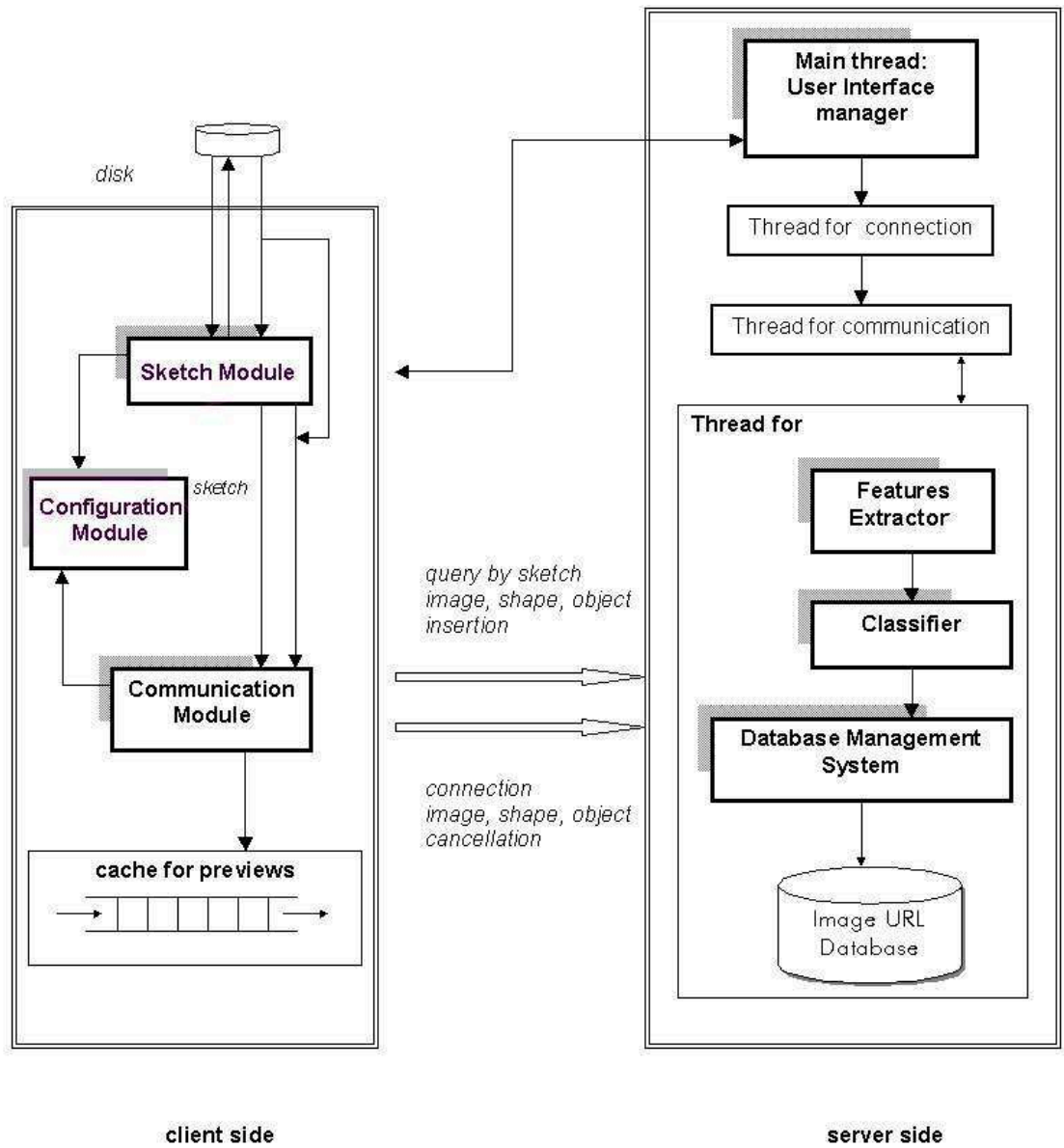

Figure 9: Architecture of the prototype system.

237

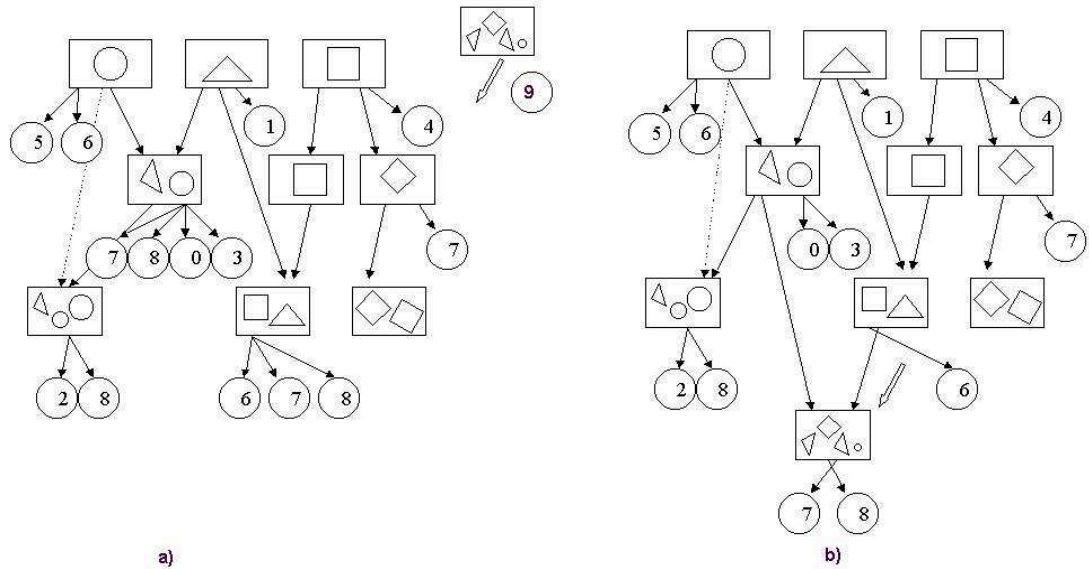

Figure 10: The process of reclassification of images when a new description is inserted: a) before insertion of description (No. 9); b) after insertion.

The communication module manages the communication with the server side, using a simple application-level protocol. The configuration module allows one to modify the parameters relative to the preview of images and shapes transferred from the server and placed in a cache managed with a FCFS policy for efficient display. The sketch module allows a user to trace basic shapes as palette items, and properly insert and modify them by varying the scale and rotation factor. The available shapes may be basic ones such as ellipse, circle, rectangle, polygons or obtained by composing the basic shapes or complex shapes defined during previous sessions of the application and inserted in the knowledge base, but also shapes extracted from segmented images.

The system keeps track of the transformations corresponding to the user's actions, and uses them in building the (internal) shape descriptions stored with the previously described syntax. The color and texture of the drawn shapes can be set according to the user requirements, as the client interface provides a color palette and the possibility to open images in JPEG format with texture content. The user can also load images from the local disk and transmit them to the server to populate the knowledge base. Finally, the user can define new objects endowing them with a textual description and insert them into the knowledge base.

The server side, which is also shown in Figure 9, is composed by concurrent threads that manage the server-side graphical interface, the connections and communications with the client applications and carry out the processing required by the client side. Obviously, the server also carries out all tasks related to the insertion of images in the knowledge base, including segmentation, feature extraction and region indexing, and allows one to properly set the various parameters involved. To this end, the server has three main subcomponents:

1. the image features extractor that contains an image segmentation module and a region data extraction one;

2. the image classifier that is composed by a classifier module and a module used in the image reclassification;

3. the database management system.

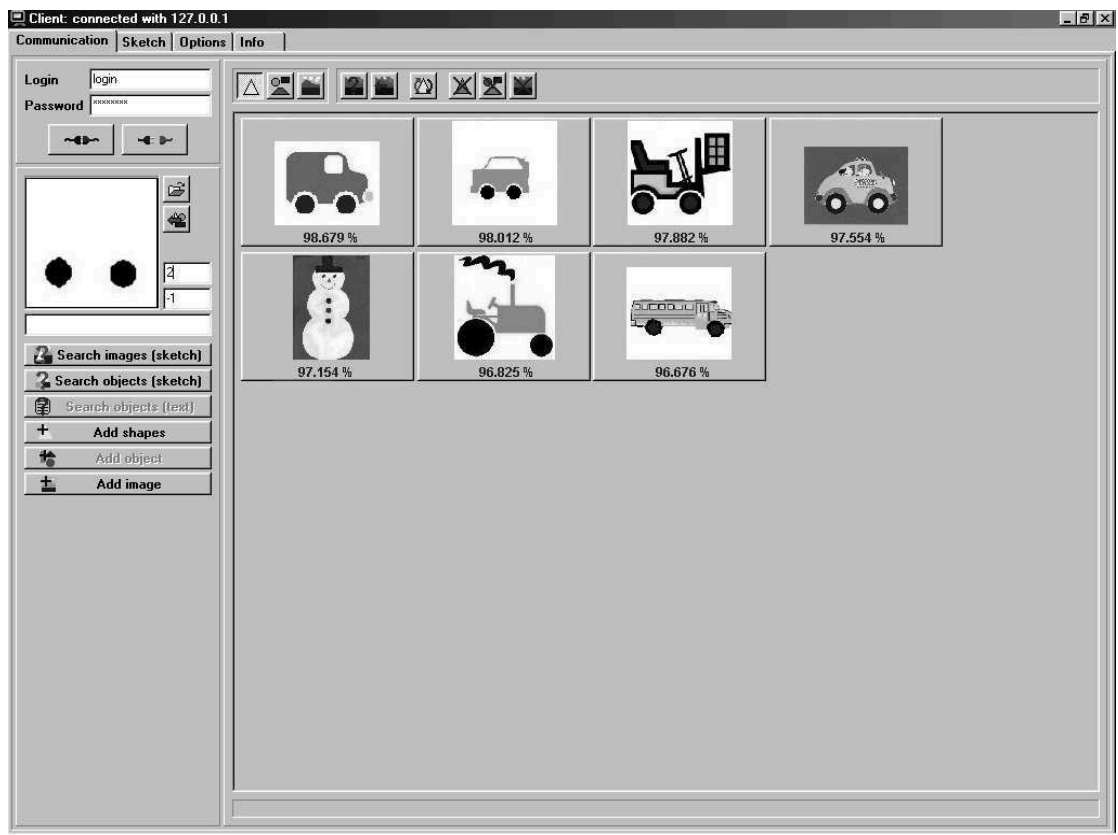

Figure 11: A query and the retrieved set of images.

The feature extractor segments and processes images to extract relevant features from each detected region, which characterize the images in the knowledge base. Image segmentation is carried out with an algorithm that starts with the extraction of relevant edges and then carries out a region growing procedure that basically merges smaller regions into larger ones according to their similarity in terms of color and texture. Detected regions obviously have to comply with some minimal heuristics. Each region has associated a description of the relevant features.

The classifier manages a graph that is used to represent and hierarchically organizes shape descriptions: basic shapes, and more complex ones obtained by combining such elementary shapes and/or by applying transformations (rotation, scaling and translation). The basic shapes have no parents, so they are at the top of the hierarchy. Images, when inserted in the knowledge base after the segmentation process, are linked to the descriptions in the structure depending on the most specific descriptions that they are able to satisfy.

The classifier module is invoked when a new description $D$ has to be inserted in the system or a new query is posed. The classifier carries out a search process in the hierarchy to find the exact position where the new description $D$ (a simple or a complex one) has to be inserted: the position is determined considering the descriptions that the new description is subsumed by. Once the position has been found, the image reclassifier compares $D$ with the images available in the database to determine those that satisfy it; all the images that verify the recognition algorithm are tied to $D$. This stage only considers the images that are tied to descriptions that are direct ancestors of $D$, as outlined in Figure 10.

As usual in Description Logics, also the query process consists of a description insertion, as both a query $Q$ and a new description $D$ are treated as prototypical images: a query $Q$ to the system is considered a new description $D$ and added to the hierarchical data structure; all images that are connected either to $Q$ or to descriptions below the query in the hierarchical structure are returned as retrieved images.

The database management module simply keeps track of images and/or pointers to images.

Using the system is a straightforward task. After logon a user can draw a sketch on the canvas combining available basic shapes, and enrich the query with color and texture content. After that the query can be posed to the server to obtain images ranked according to their similarity. Figure 11 shows a query by sketch with two circles and the retrieved set. The system correctly retrieves pictures of cars in which the two circles are recognized in the same relative positions of the sketch and represent the wheels, but also a snow man with black buttons.

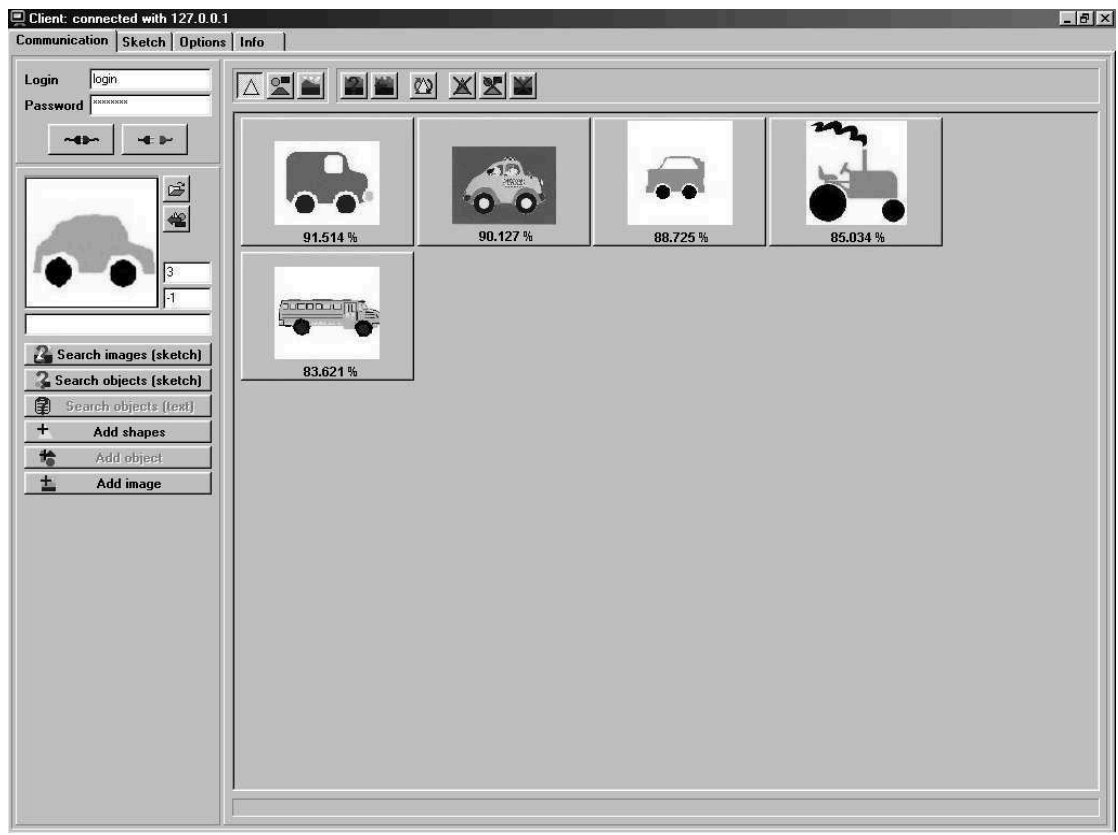

Figure 12: Downward refinement (contd.): A more detailed query, picturing a car, and the retrieved set of images.

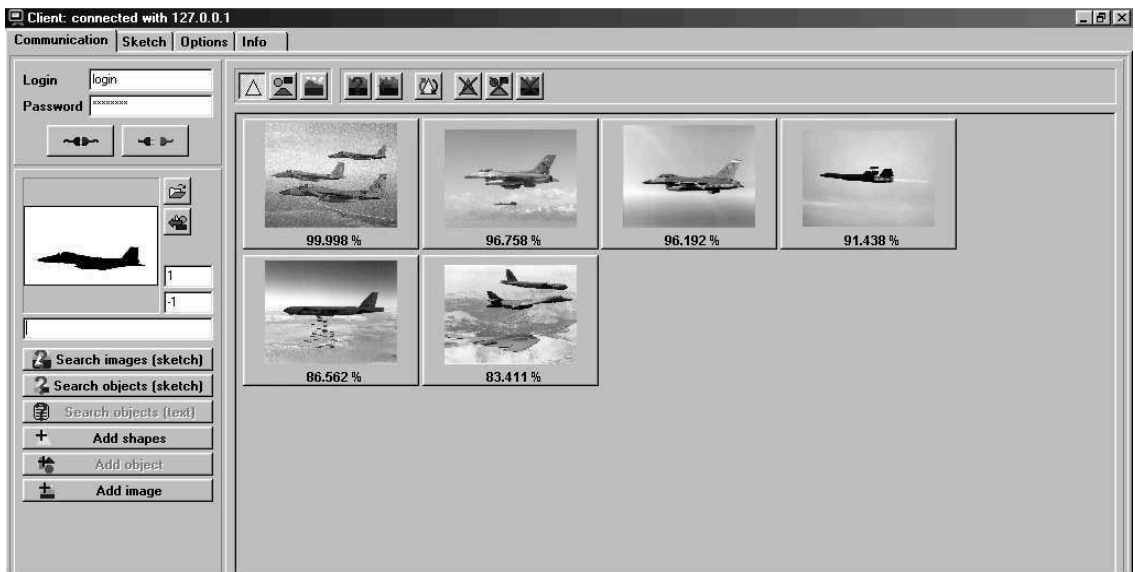

a)

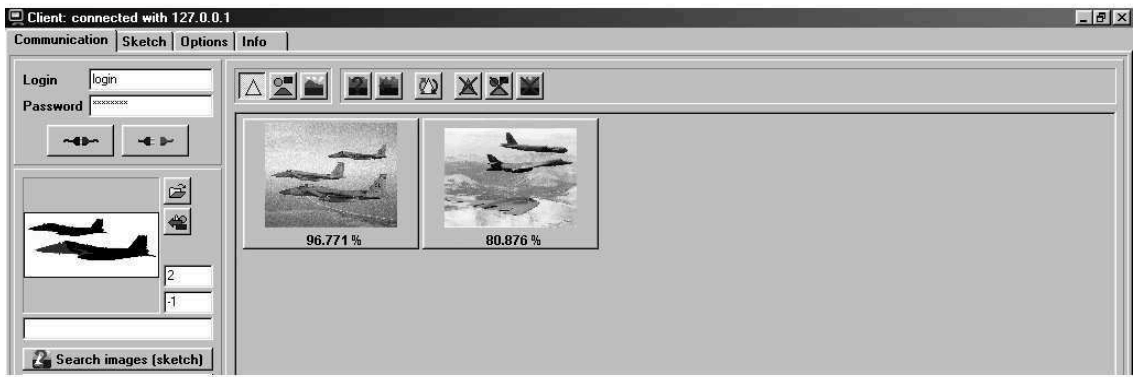

b)

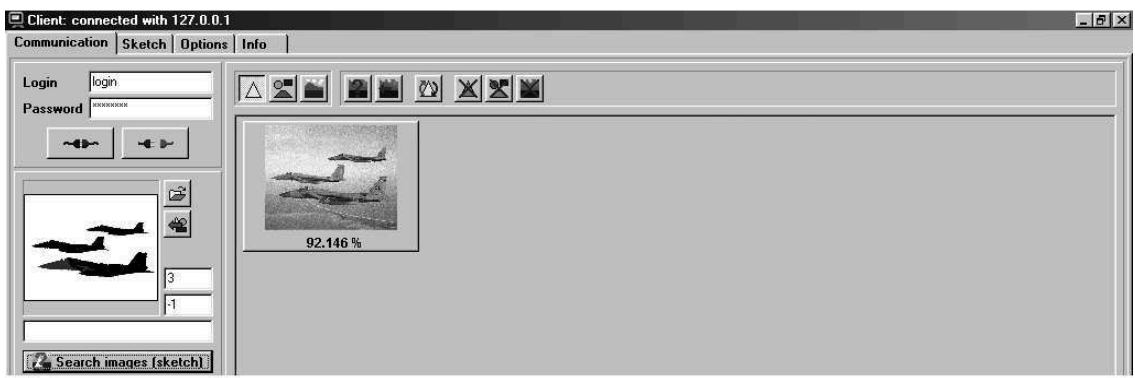

c)

Figure 13: A Subsumption example: increasing the number of objects in the query leads to a correct reduction in the retrieved set.

The introduction of more details restricts the retrieved set: adding a chassis to the previous sketch makes the query more precise, as well as the retrieval results, as it is shown in Figure 12. This example points out how we expect a user will use the system. He/she will start with a generic query with a few objects. If the number of images in the retrieved set is still too large, he/she will increase the number of details obtaining a downward refinement.

Notice that the presence of regions/objects not included in the query is obviously accepted but not the lack of a region that was explicitly introduced in the query. The idea underlying this approach is that there is an enormous amount of available images, and at the current stage of research and technology no system can always ensure a complete recognition; yet we believe that the focus should be on reducing false positives, accepting without much concern a higher ratio of false negatives. This basically means increasing precision, even at the cost of a possibly lower recall. In other words we believe it is preferable for a user looking for an image containing a yellow car, *e.g.*, using the sketch in Figure 12, that he/she receives as result of the query a limited subset of images containing almost for sure a yellow car, than a large amount of images containing cars, but also several images with no cars at all.

Subsumption is another distinguishing feature of our system. Figure 13 shows queries composed of basic shapes that have been obtained by segmentation of an image picturing aircrafts, *i.e.*, the aircraft is now a basic shape for the system. Here, to better emphasize the example, only shape and position contribute to the similarity value. The process of subsumption is clearly highlighted: a query with just a single aircraft retrieves images with one aircraft, but also with more than one aircraft. Adding other aircrafts in the graphical query correctly reduces the retrieved set. The example also points out that the system is able to correctly deal with the presence of more than one instance of an object in images, which is not possible in the approaches by Gudivada and Raghavan (1995) and Gudivada (1998). On the negative side it has to be noticed that the system did not recognize the presence of a third aircraft (indeed a strange one, the B2-Spirit) in the second image of Figure 13-b), which was not segmented at all and considered part of the background.

The ability of the system to retrieve complex objects also in images with several other different objects, that is with no "main shapes", can be anyway seen in Figure 14. Here a real image is directly submitted as query. Notice that in this case the system has to carry out the segmentation process on the fly, and detect the composing shapes.

## 6. Experiments and Results

In order to assess the performance of the proposed approach and of the system implementing it, we have carried out an extensive set of experiments on a test dataset of images. It is well known that evaluating performances of an image retrieval system is difficult because of lack of ground truth measures. To ease the possibility of a comparison, we adopted the approach first proposed by Gudivada and Raghavan (1995). The experimental framework is hence largely based on the one proposed there, which relies on a comparison of the system performances versus the judgement of human experts.

It should be noticed that in that work test images were iconic images, which were classified only in terms of spatial relationships between icons; in our experiments images are real and classification has been carried out on all image features, including color, texture, shape, scale, orientation and spatial relationships.

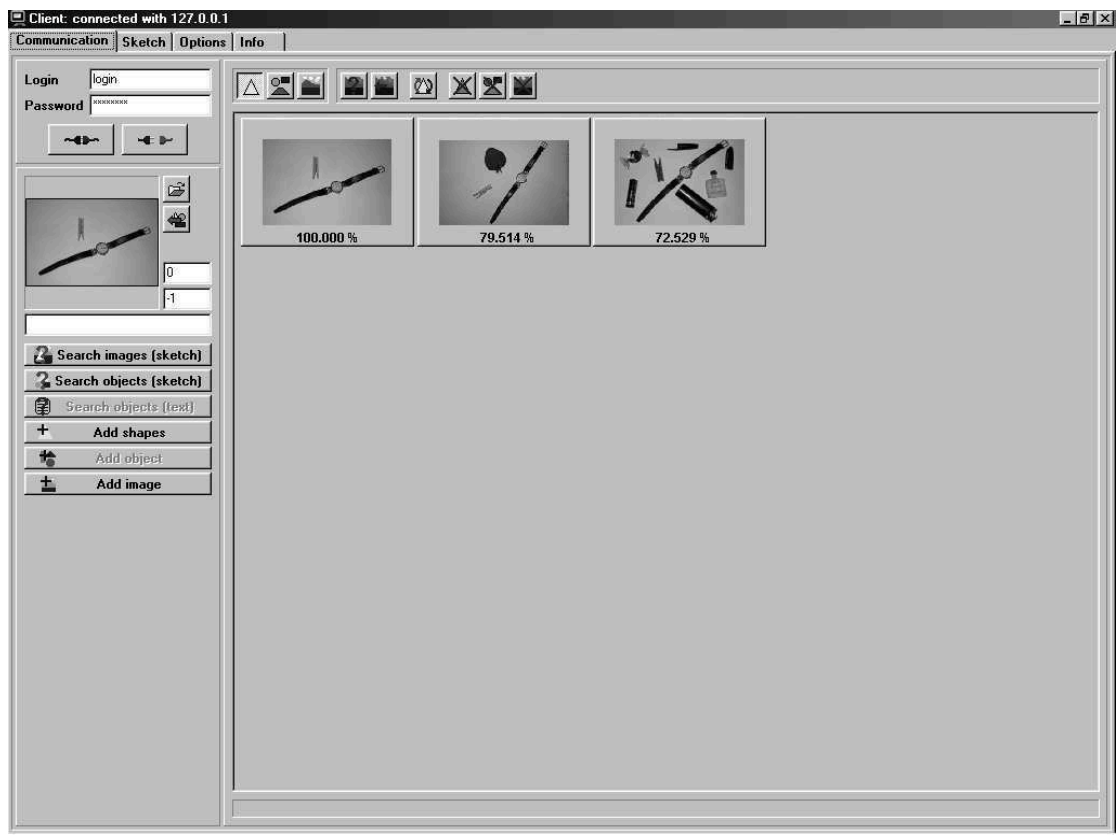

Figure 14: A query by example and retrieved images.

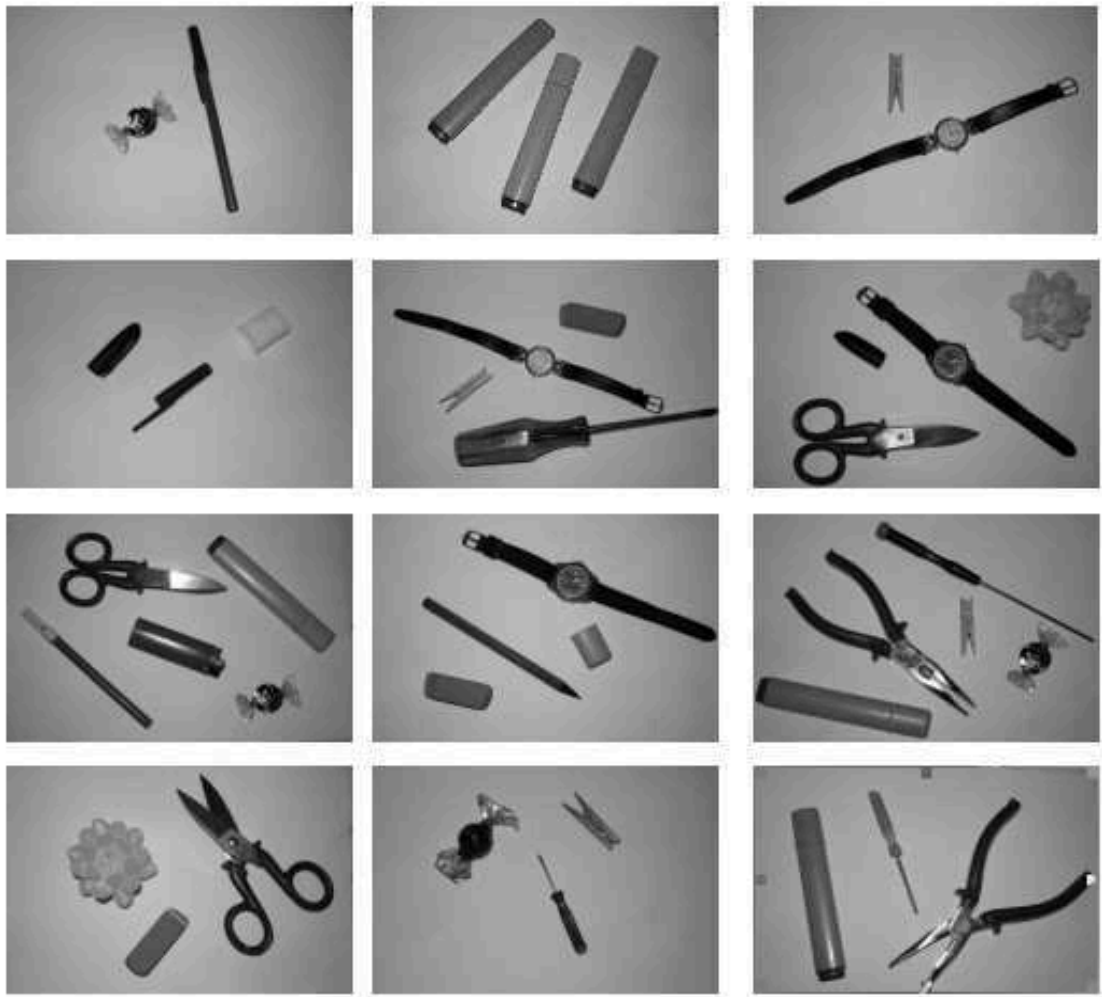

Figure 15: A sample of the images used in the experiments.

The test data set consists of a collection of 93 images; a sample of them is shown in Figure 15, while the complete set is available at URL:

    http://www-ictserv.poliba.it/disciascio/jair_images.htm.

Images have been acquired using a digital camera, combining 18 objects, either simple objects (*i.e.*, a single shape) or composite ones, of variable size and color. All images had size 1080 × 720 pixels, 24 bits/pixel. It should be noticed that actually there were more than 18 different objects, but we considered very similar variants of an object, *e.g.*, two pens with a different color, as a single test object.

We selected from the test data set 31 images to be used as queries. The query set formed two logical groupings.

The first one (namely queries 1 through 15 and queries 27, 30 and 31) had as primary objective testing the performance of the system using as query single objects composed by various shapes. That is, assessing the ability of the system to detect and retrieve images containing the same object, or objects similar to the query.

The query images in the second group (remaining images in the test data set) pictured two or more objects and they were chosen to assess the ability of the system to detect and retrieve images according to spatial relationships existing between the objects in the query.

Obviously the difference between queries containing single objects composed by several shapes, and queries containing two or more objects, is just a cognitive one: for our system all queries are composite shapes. However, we observed that performances changed for the two groupings.

We then separately asked five volunteers to classify in decreasing order, according to their judgment, the 93 images based on their similarity to each image of the selected query set. The volunteers had never used the system and they were only briefly instructed that rank orderings had to be based on the degree of conformance of the database images with the query images. They were allowed to group images when considered equivalent, and for each query, to discard images that were judged wholly dissimilar from the query.

Having obtained five classifications, which were not univocal, we created the final ranking merging the previous similarity rankings according to a minimum ranking criterion. The final ranking of each image with respect to a query was determined as the minimum one among the five available.

As an example consider the classification of Query *nr*.1, which is shown in Table 1. Notice that images grouped together in the same cell have been given the same relevance. Here Image 2 was ranked in third position by users 1,4, and 5, but as users 2 and 3 ranked it in fourth position, it was finally ranked in position four. Notice that for image 24 the same criterion leads to its withdrawal from ranked images. This approach limits the weight that images badly classified by single users have on the final ranking.

Then we submitted the same set of 31 queries to the system, whose knowledge base was loaded only with the 93 images of the test set.

The behavior of the system obviously depends on the configuration parameters, which determine the relevance of the various features involved in the similarity computation. The configuration parameters fed to the system were experimentally determined on a test bed of approximately 500 images before starting the test phase. They are shown in Table 2. The parameters reported here are described in the Appendix. Notice that, dealing with well-defined objects, we gave an higher relevance to shape and spatial features and a reduced relevance to scale, rotation, color and texture.

| user | ranking | | | | |
|---|---|---|---|---|---|
| | 1st | 2nd | 3rd | 4th | 5th |
| 1 | 1 | 44, 88 | 2, 3, 68, 80 | 26 | 24 |
| 2 | 1 | 44, 88 | 3, 68, 80 | 2, 26 | |
| 3 | 1 | 44, 88 | 3, 68, 80 | 2, 26 | |
| 4 | 1 | 44, 88 | 2, 3, 68, 80 | 26 | 24 |
| 5 | 1 | 44, 88 | 2, 3, 68, 80 | 24 26 | |
| **final** | 1 | 44, 88 | 3, 68, 80 | 2, 26 | |

Table 1: Users rankings for query $nr.1$

| Parameter | Value |
|---|---|
| Fourier descriptors threshold | 0.98 |
| Circular symmetry threshold | 0.99 |
| Spatial similarity threshold | 0.30 |
| Symmetry maxima threshold | 0.10 |
| Spatial similarity weight $\alpha$ | 0.30 |
| Spatial similarity sensitivity $fx$ | 90.0 |
| spatial similarity sensitivity $fy$ | 0.40 |
| shape similarity weight $\beta$ | 0.30 |
| shape similarity sensitivity $fx$ | 0.005 |
| shape similarity sensitivity $fy$ | 0.20 |
| color similarity weight $\gamma$ | 0.11 |
| color similarity sensitivity $fx$ | 110.0 |
| color similarity sensitivity $fy$ | 0.40 |
| rotation similarity weight $\delta$ | 0.11 |
| rotation similarity sensitivity $fx$ | 90.0 |
| rotation similarity sensitivity $fy$ | 0.40 |
| texture similarity weight $\epsilon$ | 0.07 |
| texture similarity sensitivity $fx$ | 110.0 |
| texture similarity sensitivity $fy$ | 0.40 |
| scale similarity weight $\eta$ | 0.11 |
| scale similarity sensitivity $fx$ | 0.50 |
| scale similarity sensitivity $fy$ | 0.40 |
| global similarity threshold | 0.70 |

Table 2: Configuration parameters, grouped by feature type.

The resulting classification gave us what was called a system-provided ranking. We then adopted the $R_{norm}$ as quality measure of the retrieval effectiveness. $R_{norm}$ has been first introduced in the LIVE-Project (Bollmann, Jochum, Reiner, Weissmann, & Zuse, 1985) for the evaluation of textual information retrieval systems and it has been used in the experiments of the above referenced paper by Gudivada and Raghavan. To make the paper self-contained we recall here how $R_{norm}$ is defined.

Let $G$ be a finite set of images with a user-defined preference relation $\geq$ that is complete and transitive. Let $\Delta^{usr}$ be the rank ordering of $G$ induced by the user preference relation. Also, let $\Delta^{sys}$ be a system-provided ranking. The formulation of $R_{norm}$ is:

$$R_{norm}(\Delta^{sys}) = \frac{1}{2} \cdot (1 + \frac{S^+ - S^-}{S^+_{max}})$$

where $S^+$ is the number of image pairs where a better image is ranked by the system ahead of a worse one; $S^-$ is the number of pairs where a worse image is ranked ahead of a better one and $S^+_{max}$ is the maximum possible number of $S^+$. It should be noticed that the calculation of $S^+$, $S^-$, and $S^{max}$ is based on the ranking of image pairs in $\Delta^{sys}$ relative to the ranking of corresponding image pairs in $\Delta^{usr}$.

| Query $nr.$ | Image $nr.$ | $R_{norm}$ |
|---|---|---|
| †1 | 1 | 0.92 |
| †2 | 2 | 0.92 |
| †3 | 3 | 0.93 |
| †4 | 4 | 0.95 |
| †5 | 5 | 0.99 |
| †6 | 6 | 0.94 |
| †7 | 7 | 0.93 |
| †8 | 10 | 0.93 |
| †9 | 11 | 0.95 |
| †10 | 12 | 0.74 |
| †11 | 13 | 0.60 |
| †12 | 14 | 0.84 |
| †13 | 15 | 0.83 |
| †14 | 18 | 0.99 |
| †15 | 20 | 0.91 |
| 16 | 25 | 0.89 |
| 17 | 26 | 0.80 |
| 18 | 27 | 1.00 |
| 19 | 28 | 0.74 |
| 20 | 31 | 1.00 |
| 21 | 33 | 1.00 |
| 22 | 34 | 0.99 |
| 23 | 35 | 0.91 |
| 24 | 36 | 0.89 |
| 25 | 37 | 1.00 |
| 26 | 39 | 0.99 |
| †27 | 41 | 0.93 |
| 28 | 42 | 0.98 |
| 29 | 50 | 1.00 |
| †30 | 78 | 0.88 |
| †31 | 79 | 1.00 |
| **Average $R_{norm}$** | | **0.92** |

Table 3: $R_{norm}$ values. (†*indicates single-object queries*)

$R_{norm}$ values are in the range [0,1]; a value of 1 corresponds to a system-provided ordering of the database images that is either identical to the one provided by the human experts or has a higher degree of resolution, lower values correspond to a proportional disagreement between the two.

Table 3 shows results for each query and the final average $R_{norm}$=0.92. Taking a closer look at results, for the first group of queries (single compound objects) the average value was $R_{norm}$=0.90, and $R_{norm}$=0.94 for the second grouping (various compound objects). (The complete set of result for users' ranking and system ranking is available in the online appendix).

As a comparison, the average $R_{norm}$ resulted 0.98 in the system presented by Gudivada and Raghavan (1995), where 24 iconic images were used both as queries and database images, and similarity was computed only on spatial relationships between icons. We remark here that our system works on real images and computes similarity on several image features, and we believe that results prove the ability of the system to catch to a good extent the users information need, and make refined distinctions between images when searching for composite shapes. Furthermore, our algorithm is able to correctly deal with the presence of more than one instance of an object in images, which is not possible in other approaches (Gudivada, 1998). It is also noteworthy that, though the parameters setting has been the object of several experiments, it cannot be considered optimal yet, and we believe that there is room for further improvement in the system performance, as it is also pointed out in the following paragraph.

Obviously the system can fail when segmentation does not provide accurate enough results. Figure 16 shows results for Query 11, which was the one with the worst $R_{norm}$. Here the system not only did not retrieve all images users had considered relevant, but more important wrongly confused the sugar-drop with a wrist-watch, which resulted in a false positive. As a matter of fact in various images the sweet-drops resulted not properly segmented. Nevertheless, highly relevant images were successfully retrieved and the wrongly retrieved one was slightly above the selection threshold.

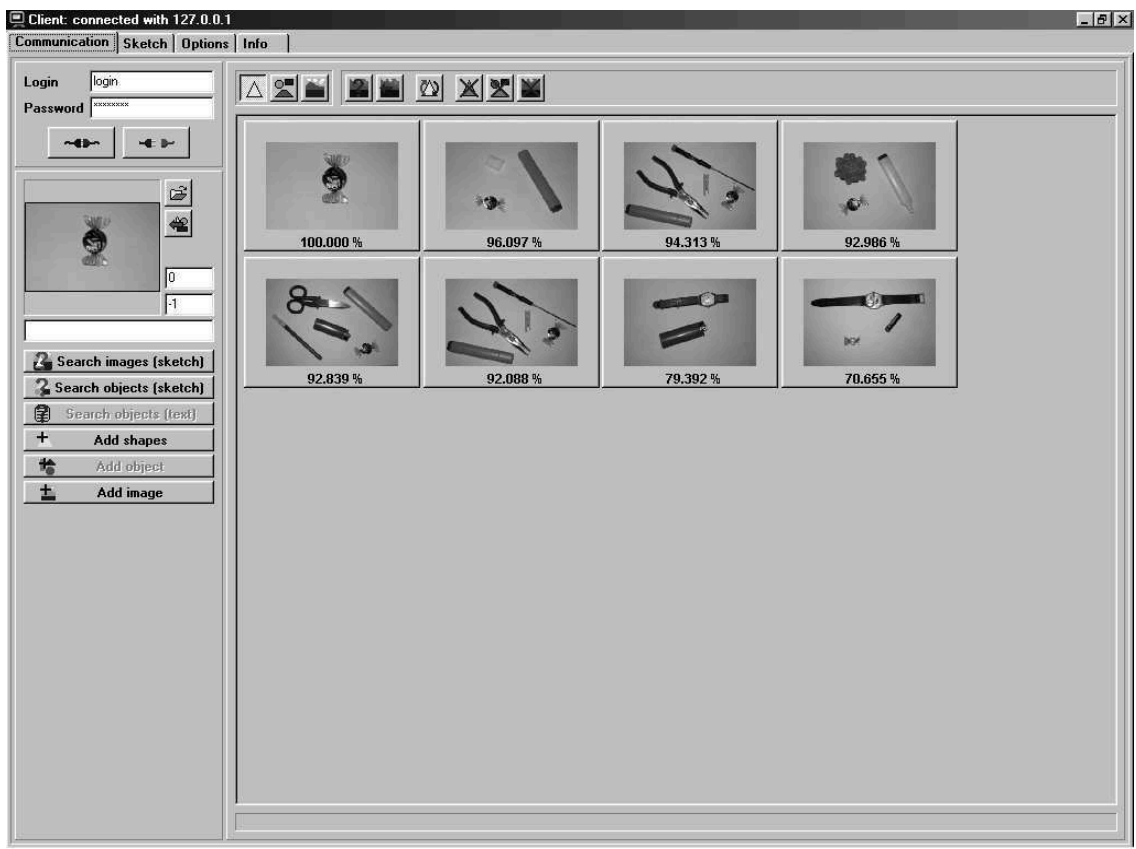

Figure 16: Query results for query 11, which had the lowest $R_{norm}=0.60$.

Another observation we made was that human users, when comparing a query with a single object, were much more driven by the color than any other feature, including the spatial positioning. This appeared in various queries and is again clearly visible using as example results for Query 11. Here users selected in the highest relevance class only images with the same color sugar-drop, and gave a lower ranking to images (with sugar-drops) with closer spatial relationships but different colors. This observation may be significant in the related field of object recognition.

A final comment. With reference to the system behavior in terms of retrieval time, we did not carry out a systematic testing, as it depends on several variables: number of images in the database, number of objects in the query, but more important depth in the hierarchy - as the search time decreases as more basic shapes are available. Limiting our analysis to the database loaded with the 93 test images, the system required on average 12 secs to answer a query, on a machine with Celeron 400 MHz CPU and 128 MB RAM running both the client and the server.

## 7. Conclusion

We proposed a Knowledge Representation approach to Image Retrieval. We started from the observation that current sketch-based image retrieval systems lack of a compositional query language — that is, they are not able to handle queries made by several shapes, where the position, orientation and size of the shapes relative to each other is meaningful.

To recover this, we proposed a language to describe composite shapes, and gave an extensional semantics to queries, in terms of sets of retrieved images. To cope with a realistic setting from the beginning, we also generalized the semantics to fuzzy membership of an image to a description. The composition of shapes is made possible by the explicit use in our language of geometric transformations (translation-rotation-scale), which we borrow form hierarchical object modeling in Computer Graphics. We believe that this is a distinguishing feature of our approach, that significantly extends standard invariant recognition of single shapes in image retrieval. The extensional semantics allows us to properly define subsumption (*i.e.*, containment) between queries.

Borrowing also from Structured Knowledge Representation, and in particular from Description Logics, we stored shape descriptions in a subsumption hierarchy. The hierarchy provides a semantic index to the images in a database. The logical semantics allowed us to define other reasoning services: the recognition of a shape arrangement in an image, the classification of an image with reference to a hierarchy of descriptions, and subsumption between descriptions. These tasks are aside, but can speed up, the main one, which is Image Retrieval.

We proved that subsumption in our simple logic can be reduced to recognition, and gave a polynomial-time algorithm to perform exact recognition. Then, for a realistic application of our setting we extended the algorithm to approximate recognition, weighting shape features (orientation, size, position), color and texture.

Using our logical approach as a formal specification, we built a prototype system using state-of-the-art technology, and set up experiments both to assess the efficacy of our proposal, and to fine tune all parameters and weights that show up in approximate retrieval. The results of our experiments, although not exhaustive, show that our approach can catch to a good extent the users information need and make refined distinctions between images when searching for composite shapes.

We believe that this proposal opens at least three directions for future research. First, the language for describing composite shapes could be enriched either with other logic-oriented connectives — *e.g.*, alternative components corresponding to an OR in compositions — or to sequences of shape arrangements, to cope with objects with internal movements in video sequence retrieval. Second, techniques from Computational Geometry could be used to optimize the algorithms for approximate retrieval, while a study in the complexity of the recognition problem for composite shapes might prove the theoretical optimality of the algorithms. Finally, large-scale experiments might prove useful in understanding the relative importance attributed by end users to the various features of a composite shape.

### Acknowledgements

We wish to thank our former students G. Gallo, M. Benedetti and L. Allegretti for their useful comments and implementations, Marco Aiello for comments on an earlier draft,

Dino Guaragnella for discussions on Fourier transforms, and an anonymous referee for constructive criticism that helped us improving the paper.

This research has been supported by the European Union, POP Regione Puglia sottomisura 7.4.1 ("SFIDA 3"), by the Italian Ministry of Education, University and Research (MIUR, ex-MURST) projects CLUSTER22 subcluster "Monitoraggio ambiente e territorio", workpackage: "Sistema informativo per il collocamento dei prodotti ortofrutticoli pugliesi" and by Italian National Council for Research (CNR), projects LAICO, DeMAnD, and "Metodi di Ragionamento Automatico nella modellazione ed analisi di dominio".

## Appendix A.

In this appendix we briefly revise the methods we used for the extraction of image features. We also describe the smoothing function $\Phi$ and the way we compute similarity for the image features that were introduced in Section 4.2.

### A.1 Extraction of Image Features

In order to deal with objects in an image, segmentation is required to obtain a partition of the image. Several segmentation algorithms have been proposed in the literature; our approach does not depend on the particular segmentation algorithm adopted. It is anyway obvious that the better the segmentation, the better our system will work. In our system we used a simple algorithm that merges edge detection and region growing.

Illustration of this technique is beyond the scope of this paper; we limit here to the description of image features computation, which assume a successful segmentation. To make the description self-contained we start defining a generic color image as $\{\overrightarrow{I}(x,y) \mid 1 \leq x \leq N_h, 1 \leq y \leq N_v\}$, where $N_h, N_v$ are the horizontal and vertical dimensions, respectively, and $\overrightarrow{I}(x,y)$ is a three-components tuple $(R, G, B)$. We assume that the image $I$ has been partitioned in $m$ regions $(r_i), i = 1, \ldots, m$ satisfying the following properties:

- $I = \bigcup (r_i), i = 1, 2, \ldots, m$

- $\forall\, i \in \{1, 2, \ldots, m\}$, $r_i$ is a nonempty and connected set

- $r_i \cap r_j = \emptyset$ iff $i \neq j$

- each region satisfies heuristic and physical requirements.

We characterize each region $r_i$ with the following attributes: shape, position, size, orientation, color and texture.

*Shape.* Given a connected region a point moving along its boundary generates a complex function defined as: $z(t) = x(t) + jy(t)$, $t = 1, \ldots, N_b$, with $N_b$ the number of boundary sample points. Following the approach proposed by Rui, She, and Huang (1996) we define the Discrete Fourier Transform (DFT) of $z(t)$ as:

$$Z(k) = \sum_{t=1}^{N_b} z(t) e^{-j\frac{2\pi t k}{N_b}} = M(k) e^{j\theta(k)}$$

with $k = 1, \ldots, N_b$.

In order to address the spatial discretization problem we compute the Fast Fourier Transform(FFT) of the boundary $z(t)$; use the first $(2N_c + 1)$ FFT coefficients to form a dense, non-uniform set of points of the boundary as:

$$z_{dense}(t) = \sum_{k=-N_c}^{N_c} Z(k) e^{-j\frac{2\pi t k}{N_b}}$$

with $t = 1, \ldots, N_{dense}$.

We then interpolate these samples to obtain uniformly spaced samples $z_{unif}(t)$, $t = 0, \ldots, N_{unif}$. We compute again the FFT of $z_{unif}(t)$ obtaining Fourier coefficients $Z_{unif}(k)$, $k = -N_c, \ldots, N_c$. The shape-feature of a region is hence characterized by a vector of $2N_c+1$ complex coefficients.

*Position and Size.* Position is determined as the region centroid computed via moment invariants (Pratt, 1991). Size is computed as the mean distance between region centroid and points on the contour.

*Orientation.* In order to quantify the orientation of each region $r_i$ we use the same Fourier representation, which stores the orientation information in the phase values. We obviously deal also with special cases when the shape of a region has more than one symmetry, *e.g.*, a rectangle or a circle. Rotational similarity between a reference shape $B$ and a given region $r_i$ can then be obtained finding maximum values via cross-correlation:

$$C(t) = \frac{1}{2N_c + 1} \sum_{k=0}^{2N_c} Z_B(k) Z_{r_i}(k) \cdot e^{j\frac{2\pi}{2N_c} k n} \text{ with } t \in 0, \ldots, 2N_c$$

*Color.* Color information of each region $r_i$ is stored, after quantization in a 112 values color space, as the mean RGB value within the region:

$$R_{r_i} = \sum_{p \in r_i} R(p) \quad G_{r_i} = \sum_{p \in r_i} G(p) \quad B_{r_i} = \sum_{p \in r_i} B(p)$$

*Texture.* We extract texture information for each region $r_i$ with a method based on the work by Pok and Liu (1999). Following this approach, we extract texture features convolving the original grey level image $I(x, y)$ with a bank of Gabor filters, having the following impulse response:

$$h(x, y) = \frac{1}{2\pi\sigma^2} \cdot e^{-\frac{x^2+y^2}{2\sigma^2}} \cdot e^{j2\pi(Ux+Vy)}$$

where $(U, V)$ represents the filter location in the frequency-domain, $\lambda$ is the central frequency, $\sigma$ is the scale factor, and $\theta$ the orientation, defined as:

$$\lambda = \sqrt{U^2 + V^2} \quad \theta = \arctan U/V$$

The processing allows to extract a 24-components feature vector, which characterizes each textured region.

## A.2 Functions for Computing Similarities

**Smoothing function $\Phi$.** In all similarity measures, we use the function $\Phi(x, fx, fy)$. The role of this function is to change a distance $x$ (in which 0 corresponds to perfect matching) to a similarity measure (in which the value 1 corresponds to perfect matching), and to "smooth" the changes of the quantity $x$, depending on two parameters $fx, fy$.

$$\Phi(x, fx, fy) = \begin{cases} fy + (1-fy) \cdot \cos(\frac{\pi x}{2 \cdot fx}) & \text{if } 0 \leq x < fx \\ fy \cdot \left[1 - \frac{\arctan[\frac{\pi \cdot (x-fx) \cdot (1-fy)}{fx \cdot fy}]}{\pi}\right] & \text{if } x > fx \end{cases}$$

where $fx > 0$ and $0 < fy < 1$.

The input data to the approximate recognition algorithm are a shape description $D$, containing $n$ components $\langle c_k, t_k, \tau_k, B_k \rangle$ and an image $I$ segmented into $m$ regions $r_1, \ldots, r_m$. The algorithm provides a measure for the approximate recognition of $D$ in $I$.

The first step of the algorithm in Section 4.2 considers all the $m$ regions the image is segmented into and all the $n$ components in the shape description D and finds — if any — all the groups of $n$ regions $r_{j(k)}$ satisfying the higher shape similarity with the shape components of D. To this purpose we compute shape similarity, based on the Fourier representation previously introduced, as vector of complex coefficients. Such measure denoted with $sim_{ss}$ is invariant with respect to rotation, scale and translation and is computed as the cosine distance between the two vectors. The similarity gives a measure in the range [0,1] assuming the higher similarity $sim_{ss} = 1$ for perfect matching.

Given the vectors $X$ and $Y$ of complex coefficients describing respectively the shape of a region $r_i$ and the shape of a component $B_k$, $X = (x_1, \ldots, x_{2N_c})$ and $Y = (y_1, \ldots, y_{2N_c})$

$$sim_{ss}(B_k, r_i) = \frac{\sum_{l=1}^{2N_c} x_l y_l}{\sqrt{\sum_{l=1}^{2N_c} x_l^2 \times \sum_{l=1}^{2N_c} y_l^2}}$$

**Shape Similarity.** The quantity $sim_{shape}$ measures the similarity between shapes in the composite shape description and the regions in the segmented image.

$$sim_{shape} = \Phi(\max_{k=1}^{n}[1 - sim_{ss}(B_k, r_{j(k)})], fx_{shape}, fy_{shape})$$

**Color Similarity.** The quantity $sim_{color}$ measures the similarity in terms of color appearance between the regions and the corresponding shapes in the composite shape description. In the following formula, $\Delta_{color}(k).R$ denotes the difference in the red color component between the $k$-th component of $D$ and the region $r_{j(k)}$, and similarly for the green and the blue color components.

$$\Delta_{color(k)} = \sqrt{[\Delta_{color}(k).R]^2 + [\Delta_{color}(k).G]^2 + [\Delta_{color}(k).B]^2}$$

Then the function $\Phi$ takes the maximum of the differences to obtain a similarity:

$$sim_{color} = \Phi(\max_{k=1}^{n}\{\Delta_{color}(k)\}, fx_{color}, fy_{color})$$

**Texture Similarity.** Finally, $sim_{texture}$ measures the similarity between the texture features in the components of $D$ and in the corresponding regions.

$\Delta_{texture}(k)$ denotes the sum of differences in the texture components between the $k$-th component of $D$ and the region $r_{j(k)}$ and dividing by the standard deviation of the elements.

$$sim_{texture} = \Phi(\max_{k=1}^{n} \Delta_{texture}(k), fx_{texture}, fy_{texture})$$